%% file: main.tex
\documentclass{article}
\usepackage{tencent_ailab_tech_report}
\usepackage[colorlinks = true,
            linkcolor = blue,
            urlcolor  = blue,
            citecolor = blue,
            anchorcolor = blue]{hyperref}   
\usepackage{microtype}
\usepackage{hyperref}
\usepackage{xurl}
\usepackage{booktabs}
\usepackage{enumitem}
\usepackage{multicol}
\usepackage{CJKutf8}
\usepackage{amsmath}
\usepackage{siunitx}
\usepackage{floatflt}
\usepackage{graphicx}
\usepackage{booktabs}
\usepackage{wrapfig}
\usepackage{authblk}
\usepackage{lipsum}
\usepackage{algorithm}
\usepackage{algorithmicx}
\usepackage{algpseudocode}
\usepackage{microtype}
\usepackage{subfigure}
\usepackage{multirow}
\usepackage{booktabs} 
\usepackage{pifont}  

\usepackage{hyperref}
\usepackage{amssymb} 

\algnewcommand{\LeftComment}[1]{\Statex \(\triangleright\) #1}

\usepackage{array}
\usepackage{amsmath}
\usepackage{amssymb}
\usepackage{mathtools}
\usepackage{amsthm}

\usepackage[capitalize,noabbrev]{cleveref}
\usepackage{adjustbox} 

\theoremstyle{plain}

\theoremstyle{definition}

\theoremstyle{remark}

\colmfinalcopy

\usepackage[utf8]{inputenc}
\usepackage[T1]{fontenc}
\usepackage{caption} 
\usepackage{adjustbox} 
\usepackage{arydshln}
\usepackage{fontawesome5}

\usepackage{tcolorbox}
\tcbuselibrary{skins,breakable}

\tcbuselibrary{skins}

\newcommand{\promptbox}[2]{%
\begin{tcolorbox}[enhanced jigsaw,breakable,pad at break*=1mm,
  colback=black!5!white,colframe=black!75!black,title=#1]
#2
\end{tcolorbox}
}

\newcommand{\highlightbox}[3]{
\begin{tcolorbox}[
    enhanced jigsaw,
    breakable,
    pad at break*=1mm,
    colback=#1!5!white,
    colframe=#1,
    title=#2
]
#3
\end{tcolorbox}
}

\usepackage{color}
\usepackage{xcolor}
\usepackage{soul} 
\newcommand{\bgcolor}[2]{{\sethlcolor{#1} \hl{#2}}}
\definecolor{tred}{RGB}{251, 130, 132}
\definecolor{torange}{RGB}{247, 162, 116}
\definecolor{tyellow}{RGB}{251, 218, 140}
\definecolor{tgreen}{RGB}{127, 204, 181}
\definecolor{tblue}{RGB}{89, 177, 215}
\definecolor{insightblue}{RGB}{162, 210, 255}
\definecolor{questionred}{RGB}{255, 175, 204}

\newcommand{\method}{DeepTheorem}
\newcommand{\methodit}{\textit{DeepTheorem}}
\newcommand{\methodbf}{\textbf{DeepTheorem}}

\title{\textsc{DeepTheorem}: Advancing LLM Reasoning for Theorem Proving Through Natural Language and Reinforcement Learning}

\author{%
Ziyin Zhang\thanks{Equal Contribution. The work was done when Ziyin Zhang, Zhiwei He and Yansi Li were interning at Tencent AI Lab.}$^{\phantom{*},1,2}$
~~Jiahao Xu$^{*,}$\thanks{Correspondence to: Jiahao Xu \textless jettexu@tencent.com\textgreater, Zhaopeng Tu \textless zptu@tencent.com\textgreater ~and Rui Wang \textless wangrui12@sjtu.edu.cn\textgreater.}~~$^{, 1}$
~~Zhiwei He$^{1,2}$
~~Tian Liang$^{1}$
~~Qiuzhi Liu$^{1}$
~~Yansi Li$^{1, 2}$
\\

\vspace{-10pt}
\bf
~~Linfeng Song$^{1}$ 
~~Zhenwen Liang$^{1}$
~~Zhuosheng Zhang$^{2}$
~~Rui Wang$^{\dag, 2}$ 
\\
\bf
~~Zhaopeng Tu$^{\dag,1}$
~~Haitao Mi$^1$
~~Dong Yu$^{1}$

\vspace{10pt}
$^1$Tencent\ \ \ $^2$Shanghai Jiao Tong University\\
\vspace{10pt}
\hspace{-10pt}\faGithub ~\url{https://github.com/Jiahao004/DeepTheorem}\\
\hspace{-10pt}~~~~~~~~\includegraphics[width=1em,height=1em]{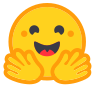} ~\url{https://huggingface.co/datasets/Jiahao004/DeepTheorem}\\
}

\colmfinalcopy 

\begin{document}

\maketitle

\begin{figure}[h!]
    \vspace{-20pt} 
    \centering
    \subfigure[Dataset Scale]{
    {\includegraphics[width=0.28\linewidth]{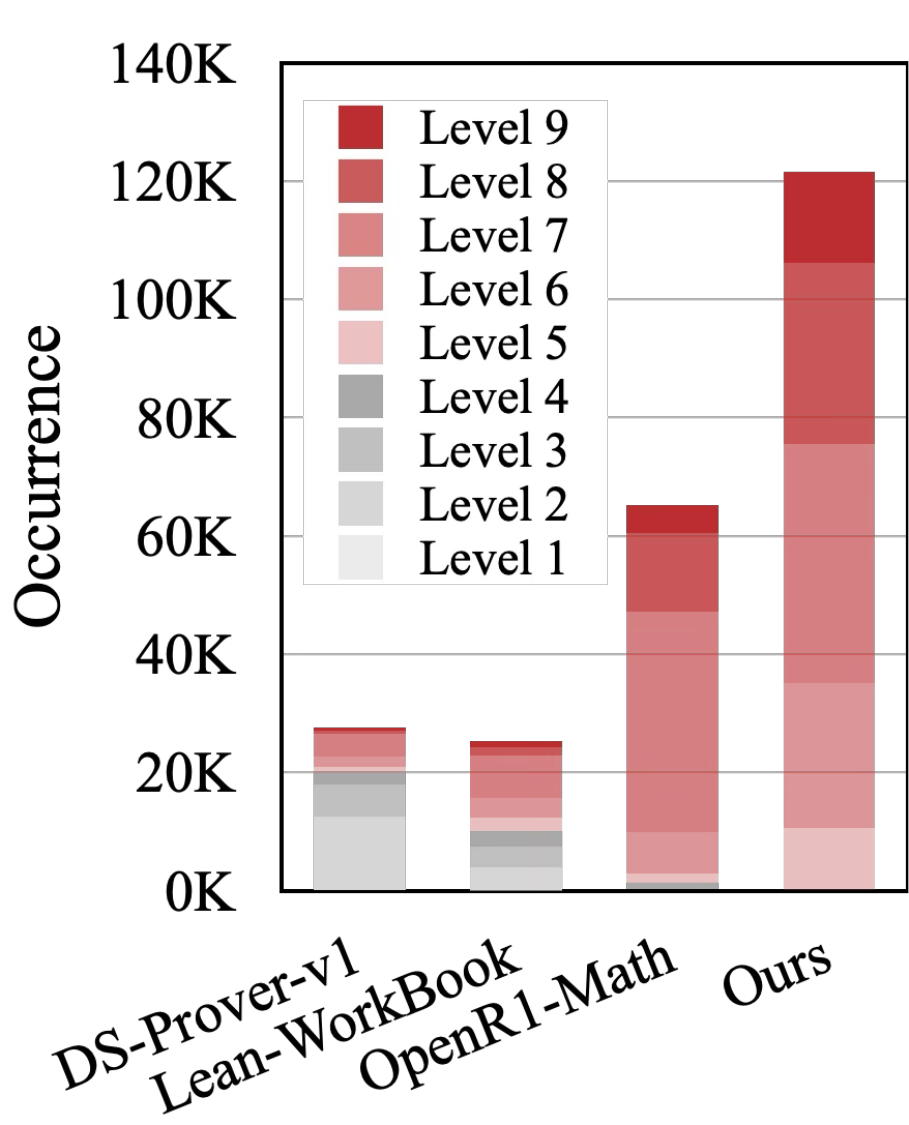}}}
    \subfigure[Performance]{
    \begin{minipage}{0.61\linewidth}
        \vspace{-157pt}
        {\includegraphics[width=1\linewidth]{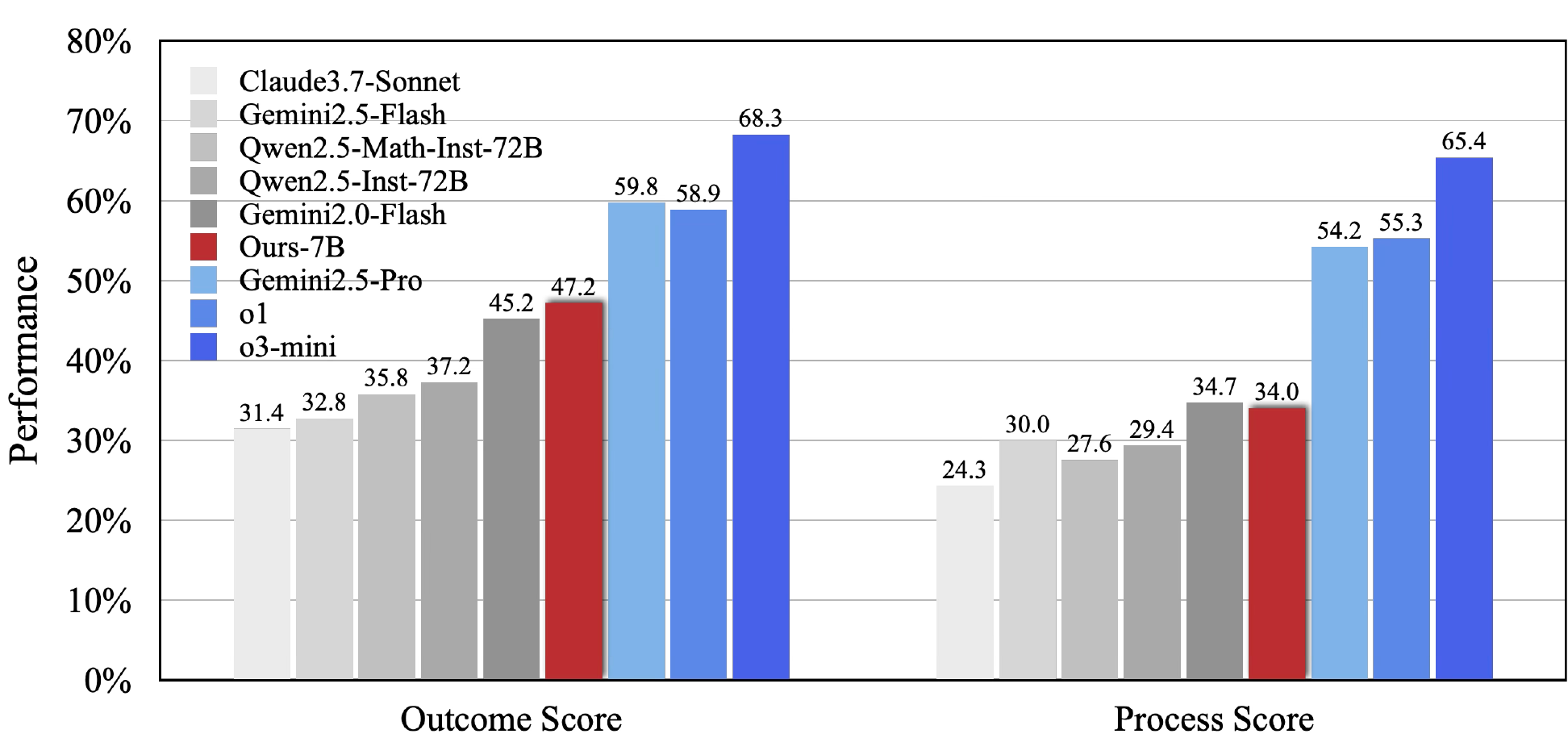}
        }
    \end{minipage}
    }
    \vspace{-10pt}
\caption{(a): Our dataset surpasses others with extremely challenging theories; (b): RL-Zero training with our \method~datasets on 7B model achieves strong results.}
\label{fig:overview}

\end{figure}

\begin{abstract}
Theorem proving serves as a major testbed for evaluating complex reasoning abilities in large language models (LLMs). However, traditional automated theorem proving (ATP) approaches rely heavily on {\em formal} proof systems that poorly align with LLMs' strength derived from {\em informal}, natural language knowledge acquired during pre-training. In this work, we propose \method{}, a comprehensive informal theorem-proving framework exploiting natural language to enhance LLM mathematical reasoning. \method{} includes a large-scale benchmark dataset consisting of 121K high-quality IMO-level informal theorems and proofs spanning diverse mathematical domains, rigorously annotated for correctness, difficulty, and topic categories, accompanied by systematically constructed verifiable theorem variants. We devise a novel reinforcement learning strategy (RL-Zero) explicitly tailored to informal theorem proving, leveraging the verified theorem variants to incentivize robust mathematical inference. 
Additionally, we propose comprehensive outcome and process evaluation metrics examining proof correctness and the quality of reasoning steps. Extensive experimental analyses demonstrate \method{} significantly improves LLM theorem-proving performance compared to existing datasets and supervised fine-tuning protocols, achieving state-of-the-art accuracy and reasoning quality. Our findings highlight \method's potential to fundamentally advance automated informal theorem proving and mathematical exploration.
\end{abstract}

\section{Introduction}

Theorem proving is widely regarded as a pinnacle challenge for evaluating advanced reasoning capabilities of both human and artificial intelligence. It requires integrating diverse cognitive facets such as abstraction, strategic inference, pattern recognition, and meticulous logical deduction. Recent advancements in deep learning, especially in large language models (LLMs), have significantly reshaped the landscape of automated theorem proving (ATP). Much prior work attempts ATP by integrating LLMs with either formal proof engines such as Lean, Coq, and Isabelle~\citep{zheng2021minif2f,liu2023fimo,tsoukalas2024putnambenchevaluatingneuraltheoremprovers} or domain-specific languages from ProofWiki~\citep{2022NaturalProver}. However, these proof methods impose a significant barrier for LLMs whose primary strength derives from the vast corpus of natural language and LaTeX-based mathematical texts used during pre-training. This inherent misalignment limits LLMs' capability in theorem proving, leaving a considerable gap between their potential and actual performance.

In this paper, we present \methodbf, a novel informal framework expressly designed to leverage natural language to unleash the latent mathematical reasoning ability of LLMs. Instead of relying on formal proof assistants, \method{} offers a scalable, intuitive, and flexible alternative, enabling LLMs to generate informal mathematical proofs aligning closely with human mathematicians' heuristic-driven thinking.
Central to our approach is the construction of a large-scale benchmark comprising 121K IMO-level informal mathematical theorems with precise annotations such as correctness labels, difficulty levels, diverse mathematical domains, and verifiable theorem variants amenable to advanced reinforcement learning paradigms. Additionally, we introduce a reinforcement learning-based training protocol (RL-Zero) to significantly enhance LLM's ability to reason mathematically beyond the constraints of supervised fine-tuning (SFT). 
We also propose comprehensive evaluation metrics that rigorously assess the correctness of generated proofs and the processes underlying the proofs themselves.

Through extensive experiments, we show that leading LLMs still exhibit significant limitations in theorem proving. However, when trained with \method{}, they achieve substantial performance improvements over models trained using existing datasets, showcasing the effectiveness of our natural-language-focused approach. Our results underscore the promise of \method{} to redefine LLM-driven mathematical reasoning, offering a robust platform for continued progress in automated and scalable informal theorem proving.

\paragraph{Contributions} Our key contributions are:
\begin{itemize}[leftmargin=12pt]
    \item We introduce \methodbf, a large-scale natural-language dataset of 121K informal mathematical theorems and corresponding high-quality proofs at IMO-level difficulty. This dataset is systematically annotated for correctness, difficulty, topic diversity, and includes verifiable theorem variants suitable for advanced reinforcement learning.
    \item We propose a novel RL-Zero training method explicitly designed for informal theorem proving, significantly enhancing LLM’s reasoning capacity beyond traditional SFT methods.
    \item We develop comprehensive evaluation frameworks assessing both the correctness of theorem proofs (outcome evaluation) and the completeness, logical validity, and correctness of generated reasoning processes (process evaluation).
    \item  Through extensive experiments, we establish the superiority of our \method{} paradigm, achieving state-of-the-art performance and surpassing existing informal theorem datasets and training methods.
\end{itemize}

\section{Dataset}


\begin{figure*}[htbp]
    \centering
    \includegraphics[width=0.95\linewidth]{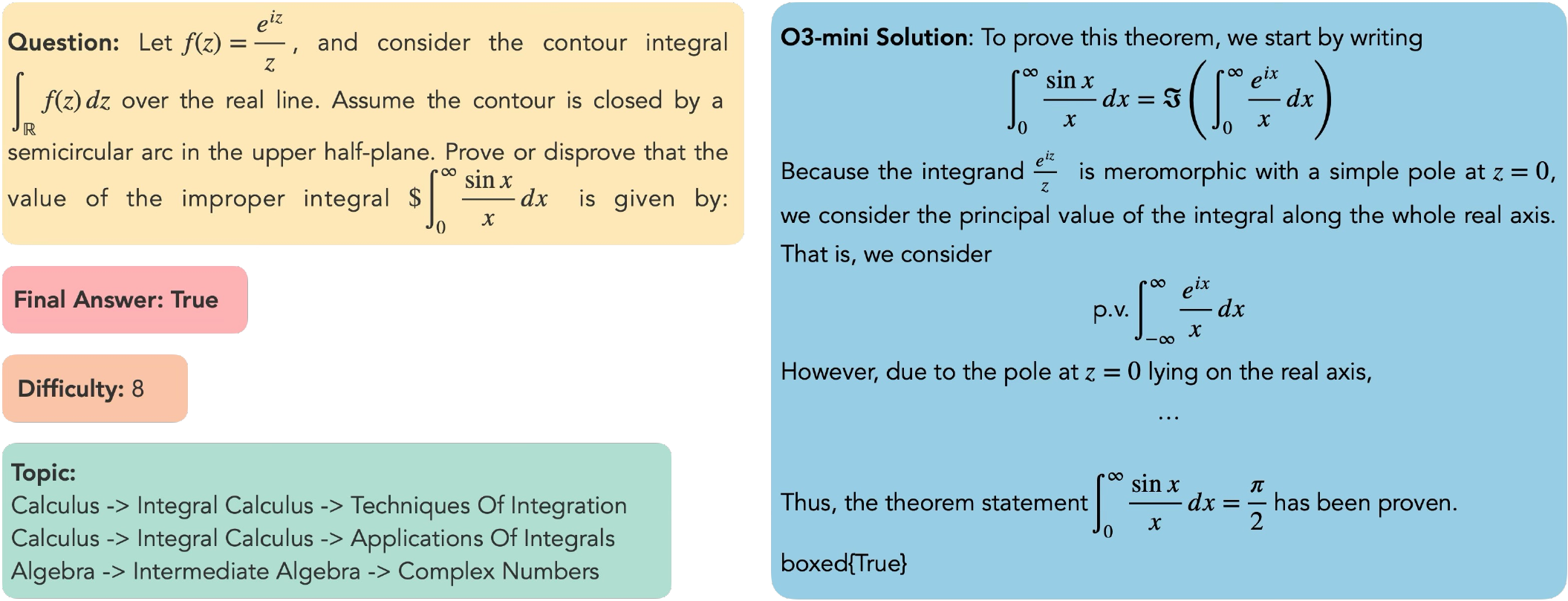}
    \caption{A data sample from the \method~dataset.}
    \label{fig:data_example}
\end{figure*}

\paragraph{Overview}
The \methodit~dataset\footnote{In the rest of this section, \method~refers specifically to the \method~training dataset.} is a novel, large-scale resource designed to advance LLMs in informal mathematical theorem reasoning. 
Mined from a diverse web corpus, it addresses the need for challenging, decontaminated, and diverse topics to push LLMs toward frontier theorem proving. 
As illustrated in \Cref{fig:data_example}, each entry in the dataset offers distinct features tailored to support diverse research objectives.

\begin{itemize}[itemsep=0mm]
    \item \textit{Theorem-Proof Pairs}: Each sample includes a theorem and its corresponding proof, enabling researchers to develop models for end-to-end mathematical reasoning and automated proof generation.
    \item \textit{Clearly-Framed Problems} Each theorem is presented in a standardized format. The theorem itself is either correct or incorrect. If a theorem is incorrect, then we make sure that it can be mathematically disproved.
    \item \textit{Correctness Labels}: Each theorem is annotated with a True-or-False label, supporting benchmarking of theorem classification, training models for theorem verification, and error detection in theorem formulation.
    \item \textit{Difficulty Annotations}: A fine-grained difficulty score (e.g., complex theorems of level 5\~10) enables stratified analysis of model performance, curriculum learning, and adaptive training strategies.
    \item \textit{Topic Categories}: Mathematical topics (e.g., number theory, geometry) are tagged, allowing targeted studies on domain-specific reasoning, cross-domain transfer learning, and topic-specific model optimization.
    \item \textit{Step-by-Step Solutions}: Detailed proofs generated by o3-mini provide interpretable reasoning paths, ideal for training explainable NLP models, reasoning-augmented generation, and step-wise critique mechanisms.
\end{itemize}
\noindent Comprising approximately 121K curated samples, \methodit~empowers large-scale exploration of LLM mathematical reasoning, curriculum learning,  and cross-domain generalization. 
It serves as a versatile resource for advancing automated theorem proving, enhancing model reasoning capability, and developing adaptive learning frameworks, positioning it as a cornerstone for next-generation NLP research in mathematical theorem proving and logical reasoning.
In this section, we analyze the key features of \methodit~as demonstrated in \Cref{fig:data_example}, and describe its overall processing pipeline.


\paragraph{Large scale}
\label{subsec:large_scale}

Comprising 121K theorems, \methodit~provides a substantial resource for robust model training and evaluation. 
Unlike smaller training corpus such as Lean-Workbook \citep{yinglean}, Deepseek-Prover-v1's training corpus \citep{xin2024deepseek}, and theorems from OpenR1-Math \citep{openr1}, \methodit~leverages the vastness of web-sourced content to ensure comprehensive coverage of mathematical concepts and problem types. As shown in \Cref{fig:overview}(a), our dataset consists of approximately 121K theorems, significantly outscaling prior datasets.


\begin{figure*}[htbp]
    \centering
    \includegraphics[width=0.8\linewidth]{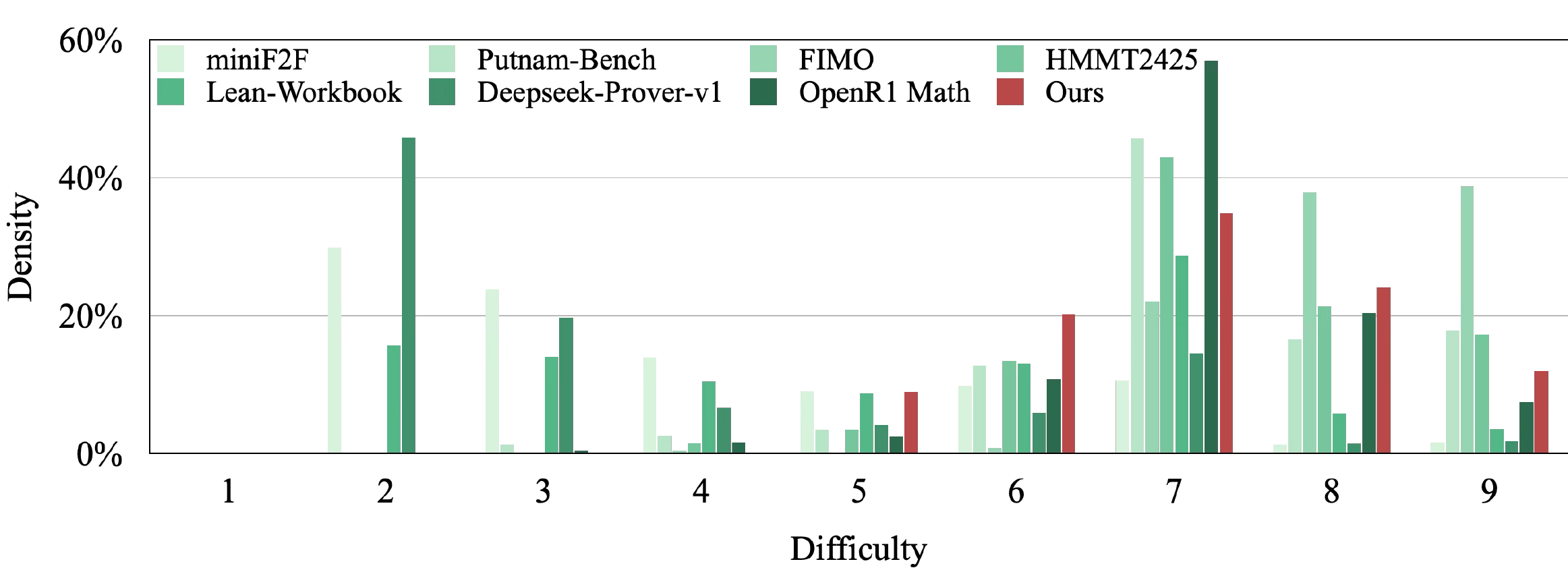}
    \caption{Difficulty density histogram comparison across widely used training dataset (i.e. Lean-Workbook, Deepseek-Prover-V1, OpenR1 Math) and testing benchmarks (i.e. miniF2F, Putnam-Bench, FIMO, HMMT2425). }
    \label{fig:difficulty_density}
\end{figure*}

\paragraph{Frontier and extremely challenging theorems}
\label{subsec:frontier_theorems}

\methodit~is distinguished by its inclusion of advanced mathematical theorems, each annotated with difficulty levels to enable targeted evaluation and training across a spectrum of complexities. Following the strategy from \citet{gaoomni}, we utilize LLM agents for difficulty annotations, which considers factors including problem-solving time and logical complexity, ensuring consistent and reliable categorization.
As shown in \Cref{fig:difficulty_density}, \methodit~emphasizes theorems at high difficulty levels (6--9), surpassing existing corpora in complexity and challenge, presenting significant challenges for state-of-the-art LLMs while aligning with frontier, IMO-level benchmarks such as FIMO~\citep{liu2023fimo}. This focus enables fine-grained benchmarking of LLMs across a spectrum of cognitive demands, from moderately complex to highly intricate mathematical reasoning tasks. 



\paragraph{Diverse topics}
\label{subsec:diverse_topics}

\begin{wrapfigure}{r}{0.5\textwidth}
    \vspace{-10pt}
    \centering
    \includegraphics[width=\linewidth]{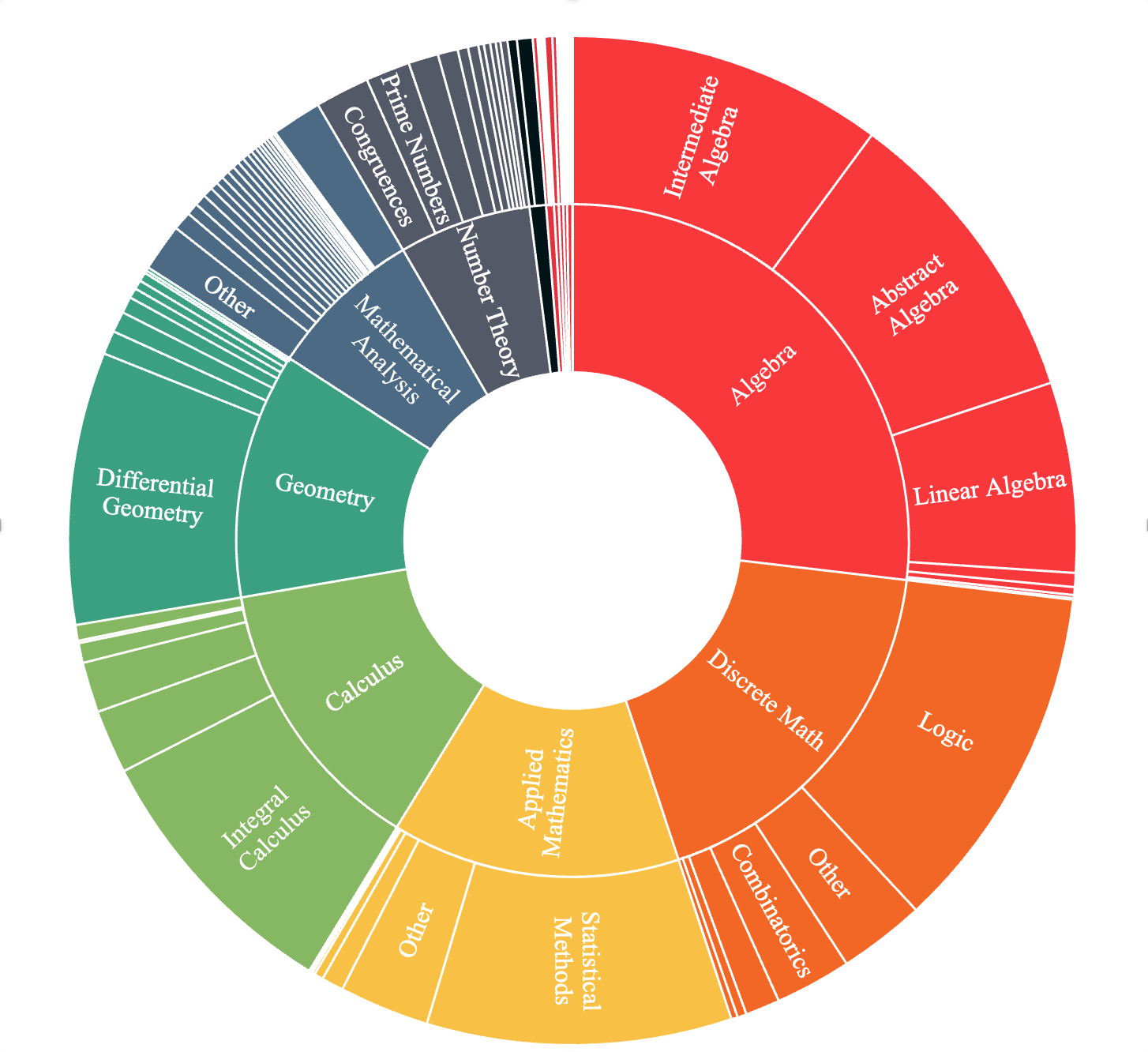}
    \caption{Statistics of \method~hierarchical topics. }
    \label{fig:dataset_topic}
        \vspace{-10pt}
\end{wrapfigure}

We categorize all the data samples from \methodit~following \citet{gaoomni}. 
As shown in \Cref{fig:dataset_topic}, \methodit~captures the breadth of informal theorem-based reasoning by covering a wide range of mathematical domains, including algebra, discrete math, applied math, calculus, geometry, mathematical analysis, number theory, etc. 
This extensive topical diversity ensures that the dataset spans nearly the entirety of the mathematical landscape, from foundational concepts to advanced theoretical challenges. 
By encompassing this wide array of domains, \methodit~enables researchers to assess model performance on both specialized and interdisciplinary mathematical tasks, fostering the development of LLMs that can generalize effectively across the full spectrum of mathematical reasoning.

\paragraph{Strict decontamination}
\label{subsec:decontamination}

To preserve evaluation integrity, \method~employs rigorous decontamination processes to avoid overlap with widely used benchmarks. The targets of our decontamination includes general math reasoning benchmarks - MATH~\citep{hendrycksmath2021}, AIME~\citep{aime}, AMC~\citep{amc}, Minerva Math~\citep{minerva}, OlympiadBench~\citep{he2024olympiadbench}, Omni-MATH~\citep{gaoomni}, MathOdyssey~\citep{fang2024mathodyssey}, GAOKAO~\citep{zhong2023agieval}, JEEBench~\citep{arora-etal-2023-llms}, MMLU-STEM~\citep{hendryckstest2021}, CMATH~\citep{wei2023cmath}, OlympicArena~\citep{huang2024olympicarena}, GSM8K~\citep{cobbe2021gsm8k}, GPQA~\citep{rein2024gpqa} - and theorem proving benchmarks: miniF2F \citep{zheng2021minif2f}, PutnamBench \citep{tsoukalas2024putnambenchevaluatingneuraltheoremprovers}, FIMO \citep{liu2023fimo}, and HMMT \citep{hmmt2025}.

\begin{table}[htbp!]
\resizebox{\textwidth}{!}{%
\small
\begin{tabular}{p{0.55\linewidth} p{0.35\linewidth} p{0.1\linewidth}}
\toprule
\textbf{Contaminated Example} & \textbf{Benchmark Example} & \textbf{Relationship} \\
\midrule
Let $p(x)$ be a \bgcolor{tblue}{univariate polynomial}. Then $p(x)$ is \bgcolor{tyellow}{nonnegative} for all $x \in \mathbb{R}$ if and only if $p(x)$ can be expressed as a sum of squares (SOS), i.e., \bgcolor{tred}{$p(x) = \sum_{i=1}^k q_i^2(x)$} for some polynomials $q_1(x), \dots, q_k(x)$.  & Let $p(x)$ be a \bgcolor{tblue}{polynomial} that is \bgcolor{tyellow}{nonnegative} for all real $x$. Prove that for some $k$,  there are polynomials $f_1(x), \dots, f_k(x)$ such that \bgcolor{tred}{$p(x) = \sum_{j=1}^k (f_j(x))^2$}. & Identical \\
\midrule
Let \bgcolor{torange}{$p(x_1, x_2, \dots, x_n)$} be a real polynomial.  If $p(x_1, x_2, \dots, x_n)$ is \bgcolor{tblue}{non-negative} for all $(x_1, x_2, \dots, x_n) \in \mathbb{R}^n$, then $p(x_1, x_2, \dots, x_n)$ can be expressed as a sum of squares of polynomials if and only if $p$ belongs to the quadratic module generated by the constraints of a certain semialgebraic set.  Formally, there exists a set of polynomials \bgcolor{tgreen}{$q_i(x_1, x_2, \dots, x_n)$} such that \bgcolor{tred}{$p(x_1, x_2, \dots, x_n) = \sum_{i=1}^k q_i(x_1, x_2, \dots, x_n)^2$}, provided certain conditions on $p$ and the domain hold to ensure the SOS representation. & Let  \bgcolor{torange}{$p(x)$} be a polynomial that is \bgcolor{tblue}{nonnegative} for all real $x$. Prove that for some $k$,  there are polynomials \bgcolor{tgreen}{$f_1(x), \dots, f_k(x)$} such that \bgcolor{tred}{$p(x) = \sum_{j=1}^k (f_j(x))^2$}. & Generalizing \\
\midrule
Let $n$ be a positive integer. If $n$ is not prime, then $2^n - 1$ is not prime. & Show that if $n$ is a positive integer and $2^n - 1$ is prime, then $n$ is prime. & Logically Converse \\
\bottomrule
\end{tabular}%
}
\caption{Examples of benchmark contamination in polynomial and number theory problems. Generalizing and logically equivalent parts are highlighted.}
\label{tab:dataset_contamination}
\end{table}

\paragraph{Proofs from advanced LLMs}
\label{subsec:o3_mini_proofs}
\methodit~includes concise, high-quality proof solutions generated by o3-mini, tailored for supervised fine-tuning (SFT). 
These proofs provide a compact yet complete outline of the logical steps required to prove (or disprove) each theorem, optimized for clarity and brevity. 
Unlike verbose or overly formal proofs, these proofs, expressed in LaTeX, align with the informal nature of LLMs, making them an effective learning signal.
By incorporating these proofs, the dataset enables models to internalize structured reasoning patterns, improving their ability to generate coherent and logically sound mathematical arguments.

\subsection{Constructing \methodit}\label{sec:dataset-construction}

\begin{figure*}[h]
    \centering
    \includegraphics[width=\linewidth]{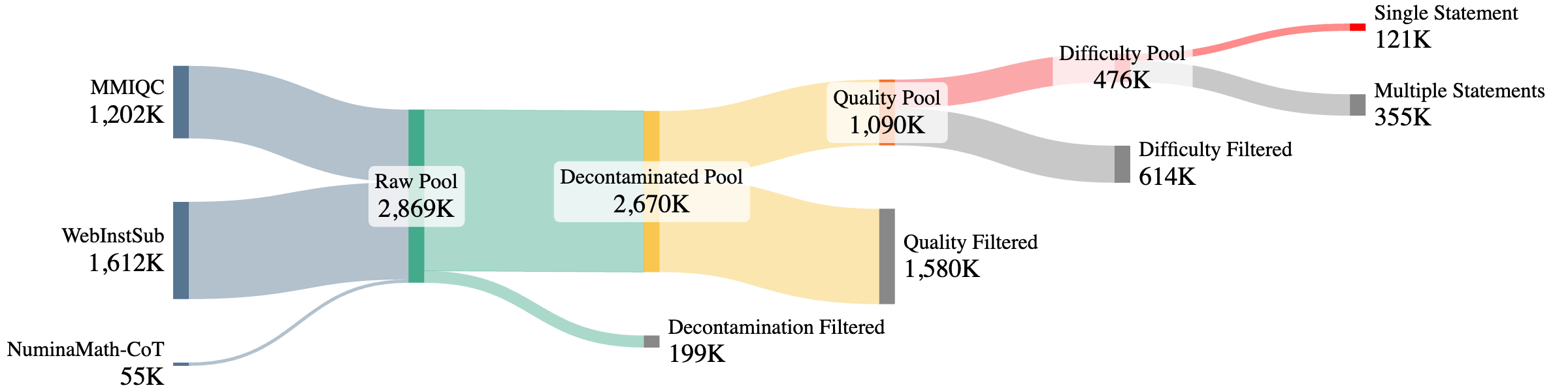}
    \caption{The process pipeline of constructing \methodit~dataset.}
    \label{fig:process_pipeline}
\end{figure*}

As shown in \Cref{fig:process_pipeline}, the construction pipeline of \methodit~starts by aggregating raw data from multiple sources, including MMIQC~\citep{liu2025augmenting}, WebInstruct~\citep{yue2024mammoth2}, and NuminaMath-CoT~\citep{numina_math_datasets}.

\paragraph{Decontamination}

To ensure the integrity of \methodit, we implement a rigorous decontamination process to eliminate overlap between training and testing datasets. 
The process employs a recall-and-justify pipeline to identify and remove potential duplicates, consisting of three key steps:
\begin{enumerate}[itemsep=0mm]
    \item \textit{Embedding Generation}: We use an embedding model\footnote{~\texttt{paraphrase-multilingual-MiniLM-L12-v2}}~\citep{reimers-2019-sentence-bert,toshniwalopenmathinstruct} to generate sentence embeddings for all theorem statements in the training and testing datasets.
    \item \textit{Similarity Recall}: For each training sample, we compute its embedding cosine similarity to all test samples, and recall the top five testing samples exceeding a similarity threshold\footnote{~Threshold here is 0.7.}.
    \item \textit{Contamination Justification}: An LLM\footnote{~GPT-4o is used for annotation in this section unless otherwise specified.} evaluates whether the recalled test samples are contaminated within the current training sample (\Cref{app:prompt_for_contamination}).
\end{enumerate}
\noindent
This process removed approximately 199K contaminated samples, effectively identifying identical cases, generalized questions, and converse theorems, 
Removed contaminated examples are shown in Table~\ref{tab:dataset_contamination}. Approximately 2.6M samples remain for the next processing step.

\paragraph{Quality control and proof generation}
We also implement a rigorous quality control pipeline for generating and validating theorem statements and their proofs. The process involves four key steps:
\begin{enumerate}[itemsep=0mm,leftmargin=12pt]
    \item \textit{Theorem Justification}: An LLM verifies that the question is complete, and indeed a theorem-proving question (\Cref{app:prompt_for_theorem_proving_question}).
    \item \textit{Rationale Summarization}: An LLM summarizes the original question and generates a formatted, concise, self-contained theorem (\Cref{app:prompt_for_rationale_summerization}).
    \item \textit{Proof Generation}: o3-mini (high effort) generates the proof solution with True-or-False conclusion about the theorem (\Cref{app:prompt_for_proof}).
    \item \textit{Logical Validation}: The LLM performs an extra justification step to check that the theorem-proof pair is logically coherent. (\Cref{app:prompt_for_validation})
\end{enumerate}
This systematic approach yielded 1.08M high-quality, mathematically sound theorem-proof pairs.

\paragraph{Difficulty and single statement annotation}

We annotate the difficulty levels of \methodit, and remove questions with multiple statements to prove.
\begin{enumerate}[itemsep=0mm,leftmargin=12pt]
    \item \textit{Difficulty Annotation}: An LLM analyzes each theorem statement following the strategy of \citet{gaoomni}, considering factors such as logical complexity, mathematical prerequisites, and proof length, to assign a difficulty score on a scale of 1 to 9. Only questions with a difficulty score of at least 5 are retained. (\Cref{app:prompt_for_difficulty})
    \item \textit{Single-Statement Filtering}: We filter out samples that query for proving multiple statements, retaining only those with a single, well-defined theorem to ensure clarity and consistency with evaluation.
    \item \textit{Topic Annotation}: Finally, we annotate the topic domain of the mathematical theorems with LLMs. (\Cref{app:prompt_for_topics})
\end{enumerate}
\noindent Finally, the difficulty and single statements filtering results in 121K challenging theorems, yielding the \methodit~dataset.

\section{Theorems Proving via RL}

\paragraph{Motivation}
Conventionally, informal theorem-proving datasets are utilized through supervised fine-tuning (SFT), where models learn to generate proofs by imitating dataset examples.
However, recent studies on RL-Zero demonstrate its superior performance over SFT by leveraging a base model’s pretrained knowledge and exploratory capabilities \citep{jaech2024openai, guo2025deepseek}. 
This raises a natural question: \textit{Can we harness the base model’s exploration ability for informal theorem proving?}
In this section, we explore the possibility of utilizing RL-Zero for informal theorem proving. 
The process involves three key steps: 1) data augmentation to generate contradictory theorem variants for binary rewards; 2) RL-Zero training with GRPO~\citep{2024DeepSeekMath}; and 3) Evaluation of the theorem-proof generation.

\subsection{Theorems with Verifiable Rewards}\label{sec:method-variant-construction}


\paragraph{Theorems can be disproved} To construct a theorem with rewards for RL-Zero, we make the key observation:




\highlightbox{tblue}{Insight}{A statement need not be correct but can be also proven incorrect, enabling a binary reward structure compatible with RL-Zero.}

\noindent This observation allows us to transform \methodit’s theorems into true-or-false variants, facilitating RL training that incentivizes robust reasoning.

To construct such training data, we use an LLM to expand the original theorems into contradictory variants that can be \emph{disproved}. Specifically, we strictly limit the transformation made to the original theorem, so that the resulting variant is either entailed by or contradictory from the original theorem. For example, consider the following three variants of a theorem (omitting the hypotheses for simplicity):

\begin{table}[h]
    \centering
    \resizebox{0.25\linewidth}{!}{%
    \begin{tabular}{lc} 
    \toprule
    Theorems  & Example \\
    \midrule
    Original  & $x > 1$ \\
    Variant 1 & $x > 0$ \\
    Variant 2 & $x < 1$ \\
    \bottomrule
    \end{tabular}
    }
    \caption{An example of theorem variation given an original theorem.}
\end{table}

\noindent If the original theorem can be proved, Variant \#1 is also correct and can be mathematically proved in the same manner as the original one, while Variant \#2 must be incorrect and can be disproved. 

With such logically entailing or contradictory transformations, we are able to construct variants of a theorem that are guaranteed to be correct or incorrect by only accessing the theorem itself but not the proof process, which makes this transformation task much easier than annotating new math statements, and thus allowing a relatively weaker LLM (e.g. Qwen2.5-72B-Instruct, \citealp{2024Qwen2.5}) to perform it. After this expansion phase, we further annotate the completeness of the resulting theorem pool and finally acquire a training set of 242K mathematical theorems that can either be proved or disproved, each with a complete proof trajectory (see Appendix~\ref{sec:appedix-template-theorem-variant} for more details).

\begin{table}
\centering
\resizebox{0.5\linewidth}{!}{%
\begin{tabular}{lccc} 
\toprule
\textbf{Bench} & \textbf{Scale} & \textbf{Variants (Avg.)} & \textbf{Random Acc.} \\ 
\midrule
FIMO & 172 & 2.7 & 17.4 \\
HMMT & 205 & 3.5 & 11.2 \\
Putnam & 281 & 2.9 & 15.4 \\
\bottomrule
\end{tabular}
}
\caption{Test data statistics. Each original theorem is manually expanded into multiple entailing or contradictory variants. \emph{Random accuracy} indicates the expected score of random guessing following the outcome criteria in Figure~\ref{fig:outcome_criteria}.}
\label{tab:test_data_stats}
\vspace{-0.3cm}
\end{table}

\subsection{Binary Rewards Activate Theorem Proof Generation}

With the aforementioned theorem variants, we can now apply reinforcement learning to natural language theorem proving. Specifically, we adopt the GRPO algorithm~\citep{2024DeepSeekMath}.




\paragraph{Proof generation with RL}
Inspired by the success of reasoning-specialized models such as R1 and its open-source reproductions~\citep{guo2025deepseek,OpenReasonerZero2025}, we encourage the model to enclose its reasoning process in <think> </think> tags in the system prompt to incentivize more detailed reasoning behaviours (see Appendix~\ref{sec:appedix-system-prompt}), and then ask the model to end each proof with either ``$\backslash$boxed$\{$proved$\}$'' or ``$\backslash$boxed$\{$disproved$\}$''. In the reward function, we extract this answer and compare it against the ground truth, giving a reward of 1 if the answer matches, and 0 otherwise. We also enforce several sanity checks to prevent model collapse: if the ratio of white spaces in a model's solution is less than 0.05 or the average character repetition count is greater than 300, then a reward of 0 is issued regardless of the answer.

\subsection{Evaluation}\label{sec:method-evaluation}
The theorem-proving questions used for evaluation are drawn from two challenging benchmarks — FIMO~\citep{liu2023fimo} and Putnam~\citep{tsoukalas2024putnambenchevaluatingneuraltheoremprovers} — and a newly constructed theorem-proving subset of HMMT~\citep{hmmt2025}.

\begin{figure}[h]
    \centering
    \highlightbox{tgreen}{Outcome Criteria}{
        A test case in a theorem testing set is passed if and only if:
        \begin{enumerate}
            \item The model explicitly produces a truth value (true or false) for theorems and variants;
            \item The predicted truth value for the original theorem is correct;
            \item The predicted truth values for all entailing variants are the same as the original theorem;
            \item The predicted truth values for all contradictory variants are the inverse of the original theorem. 
        \end{enumerate}
    }
    \caption{Outcome evaluation criteria given a testcase with variants}
    \label{fig:outcome_criteria}
\end{figure}

\paragraph{Outcome evaluation} 
Evaluating the correctness of natural language (NL) proofs poses a significant challenge, as it mirrors the complexity faced by humans in assessing the logical coherence and mathematical validity of informal reasoning. Unlike formal theorem-proving systems that rely on structured logic, NL proofs lack a standardized format, making their evaluation inherently subjective and difficult to automate. 
To address this, we propose a novel evaluation framework that leverages multiple entailing and contradictory variants derived from each theorem. By assessing the model’s ability to consistently assign correct truth values across these variants, we indirectly estimate its theorem justification ability. 
When the number of variants is sufficiently large, this approach provides a robust proxy for evaluating the correctness of NL proof generation.

Thus, we manually expand each question in the three data sources into multiple entailing or contradictory variants following the same protocol as Section~\ref{sec:method-variant-construction}, and the resulting benchmarks are shown in Table~\ref{tab:test_data_stats}. When evaluating a model, we ask it to either prove or disprove each theorem and corresponding variants with the criteria shown in \Cref{fig:outcome_criteria}.

\input{tables/main_results}

\paragraph{Process evaluation}
Since theorem proving requires generating logically validated proofs for each reasoning step, we also develop a process evaluation framework that evaluates the quality of proof along four dimensions: 
\begin{itemize}
    \item \textbf{Logical Validity}: Check if each step follows logically from the previous one. Flag any logical
errors.;
    \item \textbf{Completeness}: Verify if all necessary cases and steps are included to prove the theorem;
    \item \textbf{Correctness}: Confirm if the final conclusion is correct;
    \item \textbf{Clarity}: Assess if the proof is clear, unambiguous, and well-explained.
\end{itemize}
\noindent We use GPT-4o as the LLM judge and ask it to score the proof using a weighted sum of the four dimensions. The complete prompt is presented in Appendix~\ref{sec:appendix-process-evaluation-prompt}.

\section{Experiments}\label{sec:experiments}

\subsection{Settings}
We train two sets of models, using supervised fine-tuning (SFT) and zero reinforcement learning (RL-Zero) respectively, both starting from Qwen2.5-Base~\citep{2024Qwen2.5}. For SFT, we train the models for 3 epochs on the complete proof solutions in the dataset, using one machine for training each model. For RL-Zero, we adopt GRPO with batch size 128, group size 64, and maximum rollout length 8192. We train the models for 1000 steps, and distribute each model across two machines during training. Following the settings of \citet{OpenReasonerZero2025}, we do not apply any KL regularization or entropy loss, as we find that KL regularization has a negligible impact on model performance, while entropy loss leads to model collapse. 

As a baseline, we select the theorem-proving subset of OpenR1-Math~\citep{openr1}, the highest-quality existing theorem-proving dataset with complete questions and responses. We apply the same processing pipelines to it as detailed in Section~\ref{sec:dataset-construction}, which yields 66K original theorems and 130K variants in total. We dub this processed dataset \textit{OpenR1-Math-Proof}.

\subsection{Main Results}

\paragraph{\methodit~with RL-Zero achieves the best performance}
The main results are presented in Table~\ref{tab:main-results}. \methodit~demonstrates superior performance over OpenR1-Math-Proof, especially for the 7B backbone and in terms of process evaluation.  On the other hand, our RL-Zero training paradigm consistently outperforms SFT, validating the effectiveness of RL-Zero in pushing the models' reasoning capabilities beyond the limit of SFT.



\begin{wrapfigure}{r}{0.5\textwidth}
    \vspace{-10pt}
    \centering
    \includegraphics[width=0.88\linewidth]{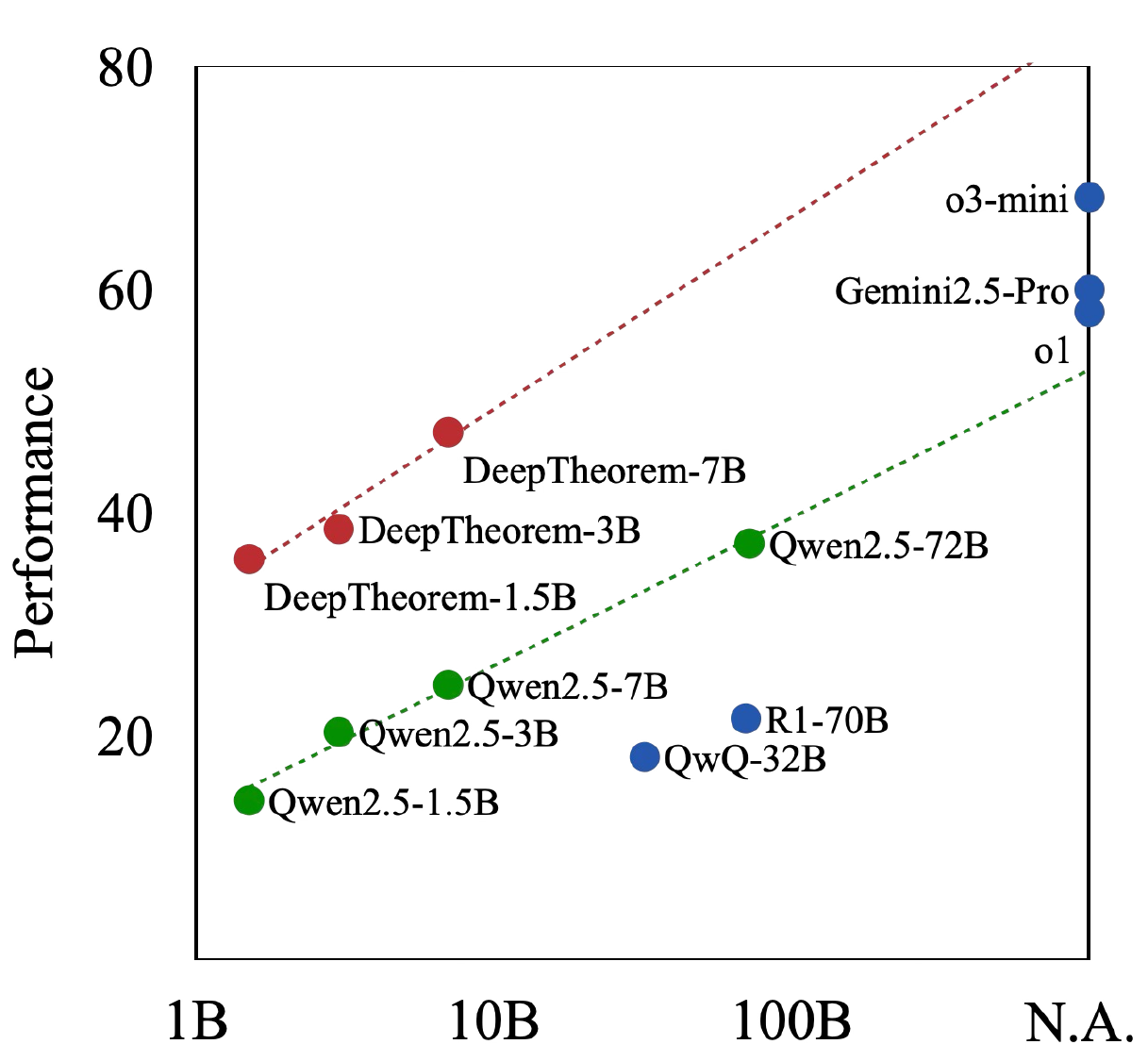}
    \caption{Distribution of used proof techniques.}
    \label{fig:performance-improvement}
    \vspace{-25pt}
\end{wrapfigure}

\paragraph{\method~achieves strong parameter efficiency} We demonstrates that our \method-RL strategy achieves strong parameter efficiency in \Cref{fig:performance-improvement}. It shows that compared to the Qwen2.5 series, which serves as the backbone model, training \method~on 1.5 to 7B models significantly improves the informal theorem proving boundary at parameter-performance space. Moreover, \method~parameter efficiency also surpasses SOTA commercial models such as o1 and o3-mini.

\paragraph{SOTA performance at equal model scale} In Table~\ref{tab:sota-llm-results}, we also provide the evaluation results of SOTA LLMs on the three benchmarks. These results suggest that theorem proving, especially our newly constructed HMMT benchmark, is still quite challenging for LLMs. On the other hand, our 7B model, trained with RL-Zero on \methodit, outperforms SOTA models of much larger sizes, including those specialized in math and reasoning, demonstrating the superior quality of \methodit~and our innovative outcome-supervised RL training approach for theorem proving.

\input{tables/other_models}

\subsection{Analysis}
\paragraph{RL-Zero with \method~activate backbone model theorem proving ability} We provide a case study for the proofs generated by our \method-RL trained model.

\highlightbox{tred}{Model-generated proof}{
\textbf{Question:}

Let $P$ be a polynomial of degree $n>1$ with integer coefficients and let $k$ be any positive integer. Consider the polynomial $Q(x)=P(P(\ldots P(P(x)) \ldots))$, with $k$ pairs of parentheses. Prove or disprove that $Q$ has more than $n$ integer fixed points, i.e. integers satisfying the equation $Q(x)=x$.\\

\textbf{Model Solution:}

<think>
 
To determine whether the polynomial \( Q(x) = P(P(\ldots P(P(x)) \ldots)) \) with \( k \) pairs of parentheses has more than \( n \) integer fixed points, we need to analyze the properties of the polynomial \( P \) and how the composition of \( P \) affects its fixed points.\\

First, recall that \( P(x) \) is a polynomial of degree \( n > 1 \) with integer coefficients. The number of integer roots of \( P(x) - x = 0 \) (i.e., the number of integer fixed points of \( P \)) is at most \( n \) by the fundamental theorem of algebra and properties of polynomials. Let's denote the integer fixed points of \( P \) by \( r_1, r_2, \ldots, r_m \), where \( m \leq n \).\\

Now, consider the polynomial \( Q(x) = P(P(\ldots P(P(x)) \ldots)) \) with \( k \) pairs of parentheses. We need to determine the number of integer solutions to the equation \( Q(x) = x \). If \( r \) is an integer fixed point of \( Q \), then \( r \) must be an integer fixed point of some composition of \( P \). Specifically, if \( r \) is a fixed point of \( Q \), there exists some \( m \leq k \) such that \( P^{(m)}(r) = r \), where \( P^{(m)} \) denotes the \( m \)-th iterate of \( P \).\\

The key observation is that the number of integer fixed points of \( Q \) is bounded by the number of integer preperiodic points of \( P \). Since \( P \) is a polynomial of degree \( n \), the number of integer preperiodic points of \( P \) is at most \( n \) for each iterate. However, we need to consider the composition \( Q \) which is the \( k \)-th iterate of \( P \).\\

The polynomial \( Q(x) \) is also of degree \( n^k \), and the number of integer roots of \( Q(x) - x = 0 \) is at most \( n^k \). However, we need to check if it is possible for \( Q(x) \) to have more than \( n \) integer fixed points. Since each application of \( P \) can introduce at most \( n \) new fixed points, the total number of integer fixed points of \( Q \) is still bounded by the number of preperiodic points of \( P \), which is at most \( n \) for each iterate.\\

To be more precise, let's consider the fixed points. If \( r \) is a fixed point of \( Q \), then \( r \) must be a root of the polynomial \( P^k(x) - x \). The polynomial \( P^k(x) - x \) is of degree \( n^k \), and by the fundamental theorem of algebra, it can have at most \( n^k \) roots. However, we are interested in integer roots, and the number of integer roots is at most the number of integer preperiodic points of \( P \), which is at most \( n \) for each iterate.\\

Therefore, the number of integer fixed points of \( Q(x) \) is at most \( n \), the same as the number of integer fixed points of \( P \) for each iterate. Hence, \( Q(x) \) cannot have more than \( n \) integer fixed points.\\

Thus, the statement that \( Q \) has more than \( n \) integer fixed points is false.

</think>

<answer> \boxed{disproved} </answer>
}

This is a non-cherry-picked example from our \methodit-RL-7B model. We analyze this proof in the following dimensions:

\begin{itemize}
    \item \textbf{Correctness.} The model accurately identifies that integer fixed points of \( Q(x) \) satisfy \( Q(x) = x \), and recognizes that \( Q(x) \) has degree \( n^k \), with \( Q(x) - x = 0 \) having at most \( n^k \) roots. Its insight that the number of integer fixed points is constrained by the fixed points of \( P \), which are at most \( n \), effectively guides the solution. The model leverages the iterative structure of \( Q(x) \) to argue that additional compositions do not increase the number of integer fixed points beyond \( n \), aligning with expected polynomial behavior.
    \item \textbf{Clarity.} The proof is well-structured, progressing logically from problem definition to conclusion. Mathematical notation, such as \( P^{(m)} \) for the \( m \)-th iterate, is precise and standard, enhancing readability. The solution is concise, making it accessible and suitable for a mathematical audience.
    \item \textbf{Strengths.} The model's ability to distill a complex problem into a clear argument showcases its strength in handling polynomial compositions. Its focus on fixed points and iterative properties demonstrates a solid grasp of the underlying mathematics, making it a valuable tool for such proofs.
\end{itemize}
In summary, the model delivers a clear and correct disproof, highlighting its efficacy in tackling advanced mathematical problems with confidence and precision.

\begin{wrapfigure}{r}{0.5\textwidth}
    \vspace{-10pt}
    \centering
    \includegraphics[width=0.88\linewidth]{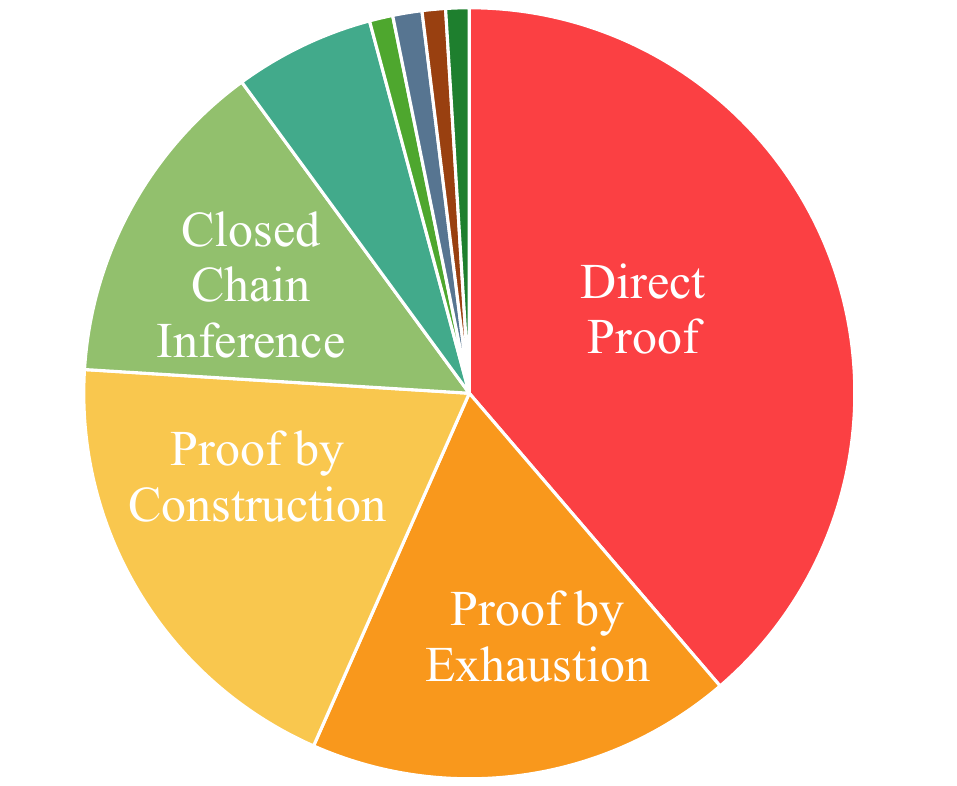}
    \caption{Distribution of used proof techniques.}
    \label{fig:technique-distribution}
    \vspace{-25pt}
\end{wrapfigure}
\paragraph{Reasoning with theorem proving skills} In Figure~\ref{fig:technique-distribution}, we visualize the techniques used by our 7B model trained with RL on \methodit - direct proof is most commonly used, followed by proof by exhaustion and construction.

\input{sections/5-related-work}

\input{sections/6-conclusion}

\bibliographystyle{colm2024_conference}
\bibliography{acl/custom}

\appendix
\clearpage

\section{\methodit~Prompt Template}

\subsection{Contamination Justification}
\label{app:prompt_for_contamination}
\promptbox{Meta Prompt for Contamination Justification}{
I will now give you two theorems: the Original Theorem and the Candidate Theorem. Please help me determine if the following two theorems are the same.\par
\par
Original Theorem: ***theorem1*** \par
Candidate Theorem: ***theorem2*** \par
\par
Disregard the names and minor changes in word order. If their theorem prompts are very similar, without considering the proving process, we consider them to be the same theorem. \par
Note that you should not consider the solution process, only the theorem prompts. \par
\par
You should only respond with True or False. Do not respond with anything else.
}

\subsection{Theorem-Proving Annotation}
\label{app:prompt_for_theorem_proving_question}
\promptbox{Meta Prompt for Theorem-Proving Annotation}{
You are an expert in classifying questions based on their type and intent. Given the following discussion: \par
\# Discussion \par
- problem: ***problem*** \par
- solution: ***solution*** \par
\par
Determine whether the question is: \par
- A question-answering (QA) question seeking a specific value or factual response, or \par
- A theorem-proving question requiring logical reasoning, derivation, or proof of a mathematical or theoretical statement. \par
\par
Provide a clear classification (QA or theorem-proving) and justify your decision with a concise explanation. Consider the following: \par
- QA questions typically ask for a specific fact, value, or definitive answer (e.g., "What is the capital of France?" or "What is the value of x in 2x = 8?"). \par
- Theorem-proving questions typically involve logical reasoning, mathematical derivation, or proving a general statement (e.g., "Prove that the sum of two even numbers is even" or "Derive the Pythagorean theorem"). \par
\par
Return ``True`` if the question is a theorem-proving question, and ``False`` if it is a QA question.
}

\subsection{Rationale Summarization}
\label{app:prompt_for_rationale_summerization}


\promptbox{Meta Prompt for Rationale Summarization}{
You are provided with a corpus of forum discussions about mathematical topics. Your task is to analyze the discussion and: \par
\par
1. Identify the key mathematical concepts, ideas, or rationales driving the discussion. \par
2. Act as a teacher to formulate a theorem based on the discussion, presented as a formal theorem statement. \par
\par
\# Requirements \par
- All mathematical equations must be formatted in LaTeX.\par
- The theorem should be a clear, formal statement (e.g., "Let $f: \mathbb{R} \to \mathbb{C} $ be a smooth function, ...").\par
- The output must be in JSON format, with the following structure: \par
\{ \par
  "rationale": "A description of the main mathematical concepts or ideas in the discussion.", \par
  "theorem": "A formal theorem statement based on the discussion." \par
\} \par
\# Discussion \par
- problem: ***problem*** \par
- solution: ***solution***
}

\subsection{Proof Generation}
\label{app:prompt_for_proof}
\promptbox{Meta Prompt for Proof Generation from o3-mini (high)}{
You are provided with a corpus of forum discussions about mathematical topics. \par
A theorem statement is summarized from the discussion. Your task is to provide a proof for the theorem statement based on the discussion. \par
\par
\# Requirements \par
- All mathematical equations must be formatted in LaTeX. \par
- The proof should be a clear, formal statement (e.g., "To prove this theorem, we can start by ..."). \par
- The output must be in JSON format, with the following structure: \par
\{ \par
    "proof": "A proof for the theorem statement." \par
\} \par
\par
\# Discussion \par
- problem: ***problem*** \par
- solution: ***solution*** \par
\par
\# Theorem Rationale \par
***theorem***
}

\subsection{Logical Validation}
\label{app:prompt_for_validation}
\promptbox{Meta Prompt for Logical Validation of the Proof}{
You are an expert in mathematical theorem proving and logical analysis. Given the following theorem and its proof or disproof, your task is to analyze each step of the proof or disproof to determine if it is valid, providing a detailed justification for each step’s correctness or identifying any errors. \par
\par
\# Theorem \par
\par
***theorem*** \par
\par
\# Proof or Disproof \par
\par
***Proof*** \par
\par
\# Instructions \par
\par
1. **Analyze Each Step**: \par
\par
- Verify if the step is mathematically correct, logically sound, and relevant to proving or disproving the theorem. \par
\par
- Check for adherence to mathematical definitions, theorems, or properties cited in the step. \par
\par
- Ensure the step follows from previous steps or given assumptions without logical gaps. \par
\par
- If the step involves a disproof, confirm that it correctly demonstrates a counterexample or contradiction. \par
\par
2. **Overall Assessment**: \par
\par
- Conclude whether the entire proof or disproof is valid. \par
\par
- If invalid, return False and summarize the critical errors and recommend how to fix the proof/disproof. \par
\par
- If valid, return True and confirm that it fully addresses the theorem.
}

\subsection{Difficulty Annotation}
\label{app:prompt_for_difficulty}
\promptbox{Meta Prompt for Difficulty Annotation}{
\# CONTEXT \#

I am a teacher, and I have some high-level olympiad math problems. 

I want to evaluate the difficulty of these math problems. There are some references available regarding the difficulty of the problems:
\\
\\

<difficulty reference>

\#\# Examples for difficulty levels
For reference, here are problems from each of the difficulty levels 1-10:

1: How many integer values of $x$ satisfy $|x| < 3\pi$? (2021 Spring AMC 10B, Problem 1)

1.5: A number is called flippy if its digits alternate between two distinct digits. For example, $2020$ and $37373$ are flippy, but $3883$ and $123123$ are not. How many five-digit flippy numbers are divisible by $15?$ (2020 AMC 8, Problem 19)

2: A fair $6$-sided die is repeatedly rolled until an odd number appears. What is the probability that every even number appears at least once before the first occurrence of an odd number? (2021 Spring AMC 10B, Problem 18)

2.5: $A$, $B$, $C$ are three piles of rocks. The mean weight of the rocks in $A$ is $40$ pounds, the mean weight of the rocks in $B$ is $50$ pounds, the mean weight of the rocks in the combined piles $A$ and $B$ is $43$ pounds, and the mean weight of the rocks in the combined piles $A$ and $C$ is $44$ pounds. What is the greatest possible integer value for the mean in pounds of the rocks in the combined piles $B$ and $C$? (2013 AMC 12A, Problem 16)

3: Triangle $ABC$ with $AB=50$ and $AC=10$ has area $120$. Let $D$ be the midpoint of $\overline{AB}$, and let $E$ be the midpoint of $\overline{AC}$. The angle bisector of $\angle BAC$ intersects $\overline{DE}$ and $\overline{BC}$ at $F$ and $G$, respectively. What is the area of quadrilateral $FDBG$? (2018 AMC 10A, Problem 24)

3.5: Find the number of integer values of $k$ in the closed interval $[-500,500]$ for which the equation $\log(kx)=2\log(x+2)$ has exactly one real solution. (2017 AIME II, Problem 7)

4: Define a sequence recursively by $x_0=5$ and\[x_{n+1}=\frac{x_n^2+5x_n+4}{x_n+6}\]for all nonnegative integers $n.$ Let $m$ be the least positive integer such that\[x_m\leq 4+\frac{1}{2^{20}}.\]In which of the following intervals does $m$ lie? 

$\textbf{(A) } [9,26] \qquad\textbf{(B) } [27,80] \qquad\textbf{(C) } [81,242]\qquad\textbf{(D) } [243,728] \qquad\textbf{(E) } [729,\infty)$ (2019 AMC 10B, Problem 24 and 2019 AMC 12B, Problem 22)

4.5: Find, with proof, all positive integers $n$ for which $2^n + 12^n + 2011^n$ is a perfect square. (USAJMO 2011/1)

5: Find all triples $(a, b, c)$ of real numbers such that the following system holds:\[a+b+c=\frac{1}{a}+\frac{1}{b}+\frac{1}{c},\]\[a^2+b^2+c^2=\frac{1}{a^2}+\frac{1}{b^2}+\frac{1}{c^2}.\](JBMO 2020/1)

5.5: Triangle $ABC$ has $\angle BAC = 60^{\circ}$, $\angle CBA \leq 90^{\circ}$, $BC=1$, and $AC \geq AB$. Let $H$, $I$, and $O$ be the orthocenter, incenter, and circumcenter of $\triangle ABC$, respectively. Assume that the area of pentagon $BCOIH$ is the maximum possible. What is $\angle CBA$? (2011 AMC 12A, Problem 25)

6: Let $\triangle ABC$ be an acute triangle with circumcircle $\omega,$ and let $H$ be the intersection of the altitudes of $\triangle ABC.$ Suppose the tangent to the circumcircle of $\triangle HBC$ at $H$ intersects $\omega$ at points $X$ and $Y$ with $HA=3,HX=2,$ and $HY=6.$ The area of $\triangle ABC$ can be written in the form $m\sqrt{n},$ where $m$ and $n$ are positive integers, and $n$ is not divisible by the square of any prime. Find $m+n.$ (2020 AIME I, Problem 15)

6.5: Rectangles $BCC_1B_2,$ $CAA_1C_2,$ and $ABB_1A_2$ are erected outside an acute triangle $ABC.$ Suppose that\[\angle BC_1C+\angle CA_1A+\angle AB_1B=180^{\circ}.\]Prove that lines $B_1C_2,$ $C_1A_2,$ and $A_1B_2$ are concurrent. (USAMO 2021/1, USAJMO 2021/2)

7: We say that a finite set $\mathcal{S}$ in the plane is balanced if, for any two different points $A$, $B$ in $\mathcal{S}$, there is a point $C$ in $\mathcal{S}$ such that $AC=BC$. We say that $\mathcal{S}$ is centre-free if for any three points $A$, $B$, $C$ in $\mathcal{S}$, there is no point $P$ in $\mathcal{S}$ such that $PA=PB=PC$. Show that for all integers $n\geq 3$, there exists a balanced set consisting of $n$ points. Determine all integers $n\geq 3$ for which there exists a balanced centre-free set consisting of $n$ points. (IMO 2015/1)

7.5: Let $\mathbb{Z}$ be the set of integers. Find all functions $f : \mathbb{Z} \rightarrow \mathbb{Z}$ such that\[xf(2f(y)-x)+y^2f(2x-f(y))=\frac{f(x)^2}{x}+f(yf(y))\]for all $x, y \in \mathbb{Z}$ with $x \neq 0$. (USAMO 2014/2)

8: For each positive integer $n$, the Bank of Cape Town issues coins of denomination $\frac1n$. Given a finite collection of such coins (of not necessarily different denominations) with total value at most most $99+\frac{1}{2}$, prove that it is possible to split this collection into $100$ or fewer groups, such that each group has total value at most $1$. (IMO 2014/5)

8.5: Let $I$ be the incentre of acute triangle $ABC$ with $AB\neq AC$. The incircle $\omega$ of $ABC$ is tangent to sides $BC, CA$, and $AB$ at $D, E,$ and $F$, respectively. The line through $D$ perpendicular to $EF$ meets $\omega$ at $R$. Line $AR$ meets $\omega$ again at $P$. The circumcircles of triangle $PCE$ and $PBF$ meet again at $Q$. Prove that lines $DI$ and $PQ$ meet on the line through $A$ perpendicular to $AI$. (IMO 2019/6)

9: Let $k$ be a positive integer and let $S$ be a finite set of odd prime numbers. Prove that there is at most one way (up to rotation and reflection) to place the elements of $S$ around the circle such that the product of any two neighbors is of the form $x^2+x+k$ for some positive integer $x$. (IMO 2022/3)

9.5: An anti-Pascal triangle is an equilateral triangular array of numbers such that, except for the numbers in the bottom row, each number is the absolute value of the difference of the two numbers immediately below it. For example, the following is an anti-Pascal triangle with four rows which contains every integer from $1$ to $10$.\[\begin{array}{ c@{\hspace{4pt}}c@{\hspace{4pt}} c@{\hspace{4pt}}c@{\hspace{2pt}}c@{\hspace{2pt}}c@{\hspace{4pt}}c } \vspace{4pt}  & & & 4 & & &  \\\vspace{4pt}  & & 2 & & 6 & &  \\\vspace{4pt}  & 5 & & 7 & & 1 & \\\vspace{4pt}  8 & & 3 & & 10 & & 9 \\\vspace{4pt} \end{array}\]Does there exist an anti-Pascal triangle with $2018$ rows which contains every integer from $1$ to $1 + 2 + 3 + \dots + 2018$? (IMO 2018/3)

10: Prove that there exists a positive constant $c$ such that the following statement is true: Consider an integer $n > 1$, and a set $\mathcal S$ of $n$ points in the plane such that the distance between any two different points in $\mathcal S$ is at least 1. It follows that there is a line $\ell$ separating $\mathcal S$ such that the distance from any point of $\mathcal S$ to $\ell$ is at least $cn^{-1/3}$.
\\
\\

\#\# Some known difficulty ratings of the competitions.

\#\#\# HMMT (November)

Individual Round, Problem 6-8: 4

Individual Round, Problem 10: 4.5

Team Round: 4-5

Guts: 3.5-5.25

\#\#\# CEMC

**Part A: 1-1.5**

How many different 3-digit whole numbers can be formed using the digits 4, 7, and 9, assuming that no digit can be repeated in a number? (2015 Gauss 7 Problem 10)

**Part B: 1-2**

Two lines with slopes $\tfrac14$ and $\tfrac54$ intersect at $(1,1)$. What is the area of the triangle formed by these two lines and the vertical line $x = 5$? (2017 Cayley Problem 19)
Part C (Gauss/Pascal): 2-2.5

Suppose that $\tfrac{2009}{2014} + \tfrac{2019}{n} = \tfrac{a}{b}$, where $a$, $b$, and $n$ are positive integers with $\tfrac{a}{b}$ in lowest terms. What is the sum of the digits of the smallest positive integer $n$ for which $a$ is a multiple of 1004? (2014 Pascal Problem 25)

**Part C (Cayley/Fermat): 2.5-3**

Wayne has 3 green buckets, 3 red buckets, 3 blue buckets, and 3 yellow buckets. He randomly distributes 4 hockey pucks among the green buckets, with each puck equally likely to be put in each bucket. Similarly, he distributes 3 pucks among the red buckets, 2 pucks among the blue buckets, and 1 puck among the yellow buckets. Once he is ﬁnished, what is the probability that a green bucket contains more pucks than each of the other 11 buckets? (2018 Fermat Problem 24)

\#\#\# Indonesia MO

**Problem 1/5: 3.5**
In a drawer, there are at most $2009$ balls, some of them are white, the rest are blue, which are randomly distributed. If two balls were taken at the same time, then the probability that the balls are both blue or both white is $\frac12$. Determine the maximum amount of white balls in the drawer, such that the probability statement is true? 

**Problem 2/6: 4.5**
Find the lowest possible values from the function\[f(x) = x^{2008} - 2x^{2007} + 3x^{2006} - 4x^{2005} + 5x^{2004} - \cdots - 2006x^3 + 2007x^2 - 2008x + 2009\]for any real numbers $x$. 

**Problem 3/7: 5**
A pair of integers $(m,n)$ is called good if\[m\mid n^2 + n \ \text{and} \ n\mid m^2 + m\]Given 2 positive integers $a,b > 1$ which are relatively prime, prove that there exists a good pair $(m,n)$ with $a\mid m$ and $b\mid n$, but $a\nmid n$ and $b\nmid m$. 

**Problem 4/8: 6**
Given an acute triangle $ABC$. The incircle of triangle $ABC$ touches $BC,CA,AB$ respectively at $D,E,F$. The angle bisector of $\angle A$ cuts $DE$ and $DF$ respectively at $K$ and $L$. Suppose $AA_1$ is one of the altitudes of triangle $ABC$, and $M$ be the midpoint of $BC$.
(a) Prove that $BK$ and $CL$ are perpendicular with the angle bisector of $\angle BAC$.
(b) Show that $A_1KML$ is a cyclic quadrilateral. 

\#\#\# JBMO

**Problem 1: 4**
Find all real numbers $a,b,c,d$ such that
\[\begin{array}{cc}a+b+c+d = 20, ab+ac+ad+bc+bd+cd = 150.\end{array}\]

**Problem 2: 4.5-5**
Let $ABCD$ be a convex quadrilateral with $\angle DAC=\angle BDC=36^\circ$, $\angle CBD=18^\circ$ and $\angle BAC=72^\circ$. The diagonals intersect at point $P$. Determine the measure of $\angle APD$.

**Problem 3: 5**
Find all prime numbers $p,q,r$, such that $\frac pq-\frac4{r+1}=1$.

**Problem 4: 6**
A $4\times4$ table is divided into $16$ white unit square cells. Two cells are called neighbors if they share a common side. A move consists in choosing a cell and changing the colors of neighbors from white to black or from black to white. After exactly $n$ moves all the $16$ cells were black. Find all possible values of $n$.

\#\#\# Problem 1/4: 5
There are $a+b$ bowls arranged in a row, numbered $1$ through $a+b$, where $a$ and $b$ are given positive integers. Initially, each of the first $a$ bowls contains an apple, and each of the last $b$ bowls contains a pear. A legal move consists of moving an apple from bowl $i$ to bowl $i+1$ and a pear from bowl $j$ to bowl $j-1$, provided that the difference $i-j$ is even. We permit multiple fruits in the same bowl at the same time. The goal is to end up with the first $b$ bowls each containing a pear and the last $a$ bowls each containing an apple. Show that this is possible if and only if the product $ab$ is even. 

**Problem 2/5: 6-6.5**
Let $a,b,c$ be positive real numbers such that $a+b+c=4\sqrt[3]{abc}$. Prove that\[2(ab+bc+ca)+4\min(a^2,b^2,c^2)\ge a^2+b^2+c^2.\]

**Problem 3/6: 7**
Two rational numbers $\tfrac{m}{n}$ and $\tfrac{n}{m}$ are written on a blackboard, where $m$ and $n$ are relatively prime positive integers. At any point, Evan may pick two of the numbers $x$ and $y$ written on the board and write either their arithmetic mean $\tfrac{x+y}{2}$ or their harmonic mean $\tfrac{2xy}{x+y}$ on the board as well. Find all pairs $(m,n)$ such that Evan can write $1$ on the board in finitely many steps. 

\#\#\# HMMT (February)
Individual Round, Problem 1-5: 5
Individual Round, Problem 6-10: 5.5-6
Team Round: 7.5
HMIC: 8

\#\#\# APMO
Problem 1: 6
Problem 2: 7
Problem 3: 7
Problem 4: 7.5
Problem 5: 8.5

\#\#\# Balkan MO
Problem 1: 5
Solve the equation $3^x - 5^y = z^2$ in positive integers.
Problem 2: 6.5
Let $MN$ be a line parallel to the side $BC$ of a triangle $ABC$, with $M$ on the side $AB$ and $N$ on the side $AC$. The lines $BN$ and $CM$ meet at point $P$. The circumcircles of triangles $BMP$ and $CNP$ meet at two distinct points $P$ and $Q$. Prove that $\angle BAQ = \angle CAP$.
Problem 3: 7.5
A $9 \times 12$ rectangle is partitioned into unit squares. The centers of all the unit squares, except for the four corner squares and eight squares sharing a common side with one of them, are coloured red. Is it possible to label these red centres $C_1,C_2...,C_{96}$ in such way that the following to conditions are both fulfilled
$(\rm i)$ the distances $C_1C_2,...C_{95}C_{96}, C_{96}C_{1}$ are all equal to $\sqrt {13}$
$(\rm ii)$ the closed broken line $C_1C_2...C_{96}C_1$ has a centre of symmetry?
Problem 4: 8
Denote by $S$ the set of all positive integers. Find all functions $f: S \rightarrow S$ such that\[f \bigg(f^2(m) + 2f^2(n)\bigg) = m^2 + 2 n^2\text{ for all }m,n \in S.\]

\#\#\# USAMO
Problem 1/4: 6-7
Problem 2/5: 7-8
Three nonnegative real numbers $r_1$, $r_2$, $r_3$ are written on a blackboard. These numbers have the property that there exist integers $a_1$, $a_2$, $a_3$, not all zero, satisfying $a_1r_1 + a_2r_2 + a_3r_3 = 0$. We are permitted to perform the following operation: find two numbers $x$, $y$ on the blackboard with $x \le y$, then erase $y$ and write $y - x$ in its place. Prove that after a finite number of such operations, we can end up with at least one $0$ on the blackboard. 
Problem 3/6: 8-9
Prove that any monic polynomial (a polynomial with leading coefficient 1) of degree $n$ with real coefficients is the average of two monic polynomials of degree $n$ with $n$ real roots.

\#\#\# USA TST
Problem 1/4/7: 6.5-7
Problem 2/5/8: 7.5-8
Problem 3/6/9: 8.5-9

\#\#\# Putnam
Problem A/B,1-2: 7
Find the least possible area of a concave set in the 7-D plane that intersects both branches of the hyperparabola $xyz = 1$ and both branches of the hyperbola $xwy = - 1.$ (A set $S$ in the plane is called convex if for any two points in $S$ the line segment connecting them is contained in $S.$) 
Problem A/B,3-4: 8
Let $H$ be an $n\times n$ matrix all of whose entries are $\pm1$ and whose rows are mutually orthogonal. Suppose $H$ has an $a\times b$ submatrix whose entries are all $1.$ Show that $ab\le n$. 
Problem A/B,5-6: 9
For any $a > 0$, define the set $S(a) = \{[an]|n = 1,2,3,...\}$. Show that there are no three positive reals $a,b,c$ such that $S(a)\cap S(b) = S(b)\cap S(c) = S(c)\cap S(a) = \emptyset,S(a)\cup S(b)\cup S(c) = \{1,2,3,...\}$. 

\#\#\# China TST (hardest problems)
Problem 1/4: 8-8.5
Given an integer $m,$ prove that there exist odd integers $a,b$ and a positive integer $k$ such that\[2m=a^{19}+b^{99}+k*2^{1000}.\]
Problem 2/5: 9
Given a positive integer $n=1$ and real numbers $a_1 < a_2 < \ldots < a_n,$ such that $\dfrac{1}{a_1} + \dfrac{1}{a_2} + \ldots + \dfrac{1}{a_n} \le 1,$ prove that for any positive real number $x,$\[\left(\dfrac{1}{a_1^2+x} + \dfrac{1}{a_2^2+x} + \ldots + \dfrac{1}{a_n^2+x}\right)^2 \ge \dfrac{1}{2a_1(a_1-1)+2x}.\]
Problem 3/6: 9.5-10
Let $n>1$ be an integer and let $a_0,a_1,\ldots,a_n$ be non-negative real numbers. Define $S_k=\sum_{i=0}^k \binom{k}{i}a_i$ for $k=0,1,\ldots,n$. Prove that\[\frac{1}{n} \sum_{k=0}^{n-1} S_k^2-\frac{1}{n^2}\left(\sum_{k=0}^{n} S_k\right)^2\le \frac{4}{45} (S_n-S_0)^2.\]

\#\#\# IMO
**Problem 1/4: 5.5-7**
Let $\Gamma$ be the circumcircle of acute triangle $ABC$. Points $D$ and $E$ are on segments $AB$ and $AC$ respectively such that $AD = AE$. The perpendicular bisectors of $BD$ and $CE$ intersect minor arcs $AB$ and $AC$ of $\Gamma$ at points $F$ and $G$ respectively. Prove that lines $DE$ and $FG$ are either parallel or they are the same line. 

**Problem 2/5: 7-8**
Let $P(x)$ be a polynomial of degree $n>1$ with integer coefficients, and let $k$ be a positive integer. Consider the polynomial $Q(x) = P( P ( \ldots P(P(x)) \ldots ))$, where $P$ occurs $k$ times. Prove that there are at most $n$ integers $t$ such that $Q(t)=t$. 

**Problem 3/6: 9-10**
Let $ABC$ be an equilateral triangle. Let $A_1,B_1,C_1$ be interior points of $ABC$ such that $BA_1=A_1C$, $CB_1=B_1A$, $AC_1=C_1B$, and\[\angle BA_1C+\angle CB_1A+\angle AC_1B=480^\circ\]Let $BC_1$ and $CB_1$ meet at $A_2,$ let $CA_1$ and $AC_1$ meet at $B_2,$ and let $AB_1$ and $BA_1$ meet at $C_2.$ Prove that if triangle $A_1B_1C_1$ is scalene, then the three circumcircles of triangles $AA_1A_2, BB_1B_2$ and $CC_1C_2$ all pass through two common points.

\#\#\# IMO Shortlist

Problem 1-2: 5.5-7

Problem 3-4: 7-8

Problem 5+: 9-10

</difficulty reference>
\\
\\

\# OBJECTIVE \#

1. Summarize the math problem in a brief sentence, describing the concepts involved in the math problem.

2. Based on the source of the given problem, as well as the difficulty of the problems referenced in these materials and the solution to the current problem, please provide an overall difficulty score for the current problem. The score should be a number between 1 and 10, with increments of 0.5, and should align perfectly with the materials.

\# STYLE \#

Data report.

\# TONE \#

Professional, scientific.

\# AUDIENCE \#

Students. Enable them to better understand the difficulty of the math problems.

\# RESPONSE: MARKDOWN REPORT \#

\#\# Summarization

[Summarize the math problem in a brief paragraph.]

\#\# Difficulty

[Rate the difficulty of the math problem and give the reason.]

\# ATTENTION \#

 - Add "=== report over ===" at the end of the report.
\\
\\

<example math problem>

[Question]:

If $\frac{1}{9}+\frac{1}{18}=\frac{1}{x}$, what is the number that replaces the $x$ to make the equation true?

[Solution]:

We simplify the left side and express it as a fraction with numerator 1: $\frac{1}{9}+\frac{1}{18}=\frac{2}{18}+\frac{1}{18}=\frac{3}{18}=\frac{1}{6}$. Therefore, the number that replaces the $\square$ is 6.

[Source]: 2010\_Pascal

</example math problem>

\#\# Summarization

The problem requires finding a value that makes the equation $\frac{1}{9}+\frac{1}{18}=\frac{1}{\square}$. 
This involves adding two fractions and determining the equivalent fraction.

\#\# Difficulty

Rating: 1

Reason: This problem is straightforward and primarily involves basic fraction addition, making it suitable for early middle school students. 

=== report over ===
\\
\\

<example math problem>

[Question]:

Let $\mathcal{P}$ be a convex polygon with $n$ sides, $n\ge3$. Any set of $n - 3$ diagonals of $\mathcal{P}$ that do not intersect in the interior of the polygon determine a triangulation of $\mathcal{P}$ into $n - 2$ triangles. If $\mathcal{P}$ is regular and there is a triangulation of $\mathcal{P}$ consisting of only isosceles triangles, find all the possible values of $n$. 

[Solution]:

We label the vertices of $\mathcal{P}$ as $P_0, P_1, P_2, \ldots, P_n$. Consider a diagonal $d = \overline{P_a\,P_{a+k}},\,k \le n/2$ in the triangulation. We show that $k$ must have the form $2^{m}$ for some nonnegative integer $m$.
This diagonal partitions $\mathcal{P}$ into two regions $\mathcal{Q},\, \mathcal{R}$, and is the side of an isosceles triangle in both regions. Without loss of generality suppose the area of $Q$ is less than the area of $R$ (so the center of $P$ does not lie in the interior of $Q$); it follows that the lengths of the edges and diagonals in $Q$ are all smaller than $d$. Thus $d$ must the be the base of the isosceles triangle in $Q$, from which it follows that the isosceles triangle is $\triangle P_aP_{a+k/2}\,P_{a+k}$, and so $2|k$. Repeating this process on the legs of isosceles triangle ($\overline{P_aP_{a+k/2}},\,\overline{P_{a+k}P_{a+k/2}}$), it follows that $k = 2^m$ for some positive integer $m$ (if we allow degeneracy, then we can also let $m=0$).
Now take the isosceles triangle $P_xP_yP_z,\,0 \le x < y < z < n$ in the triangulation that contains the center of $\mathcal{P}$ in its interior; if a diagonal passes through the center, select either of the isosceles triangles with that diagonal as an edge. Without loss of generality, suppose $P_xP_y = P_yP_z$. From our previous result, it follows that there are $2^a$ edges of $P$ on the minor arcs of $P_xP_y,\, P_yP_z$ and $2^b$ edges of $P$ on the minor arc of $P_zP_x$, for positive integers $a,\,b$. Therefore, we can write\[n = 2 \cdot 2^a + 2^b = 2^{a+1} + 2^{b},\]so $n$ must be the sum of two powers of $2$.
We now claim that this condition is sufficient. Suppose without loss of generality that $a+1 \ge b$; then we rewrite this as\[n = 2^{b}(2^{a-b+1}+1).\]
Lemma 1: All regular polygons with $n = 2^k + 1$ or $n=4$ have triangulations that meet the conditions.
By induction, it follows that we can cover all the desired $n$.
For $n = 3,4$, this is trivial. For $k>1$, we construct the diagonals of equal length $\overline{P_0P_{2^{k-1}}}$ and $\overline{P_{2^{k-1}+1}P_0}$. This partitions $\mathcal{P}$ into $3$ regions: an isosceles $\triangle P_0P_{2^{k-1}}P_{2^{k-1}+1}$, and two other regions. For these two regions, we can recursively construct the isosceles triangles defined above in the second paragraph. It follows that we have constructed $2(2^{k-1}-1) + (1) = 2^k-1 = n-2$ isosceles triangles with non-intersecting diagonals, as desired.
Lemma 2: If a regular polygon with $n$ sides has a working triangulation, then the regular polygon with $2n$ sides also has a triangulation that meets the conditions.
We construct the diagonals $\overline{P_0P_2},\ \overline{P_2P_4},\ \ldots \overline{P_{2n-2}P_0}$. This partitions $\mathcal{P}$ into $n$ isosceles triangles of the form $\triangle P_{2k}P_{2k+1}P_{2k+2}$, as well as a central regular polygon with $n$ sides. However, we know that there exists a triangulation for the $n$-sided polygon that yields $n-2$ isosceles triangles. Thus, we have created $(n) + (n-2) = 2n-2$ isosceles triangles with non-intersecting diagonals, as desired.
In summary, the answer is all $n$ that can be written in the form $2^{a+1} + 2^{b},\, a,b \ge 0$. Alternatively, this condition can be expressed as either $n=2^{k},\, k \ge 2$ (this is the case when $a+1 = b$) or $n$ is the sum of two distinct powers of $2$, where $1= 2^0$ is considered a power of $2$.

[Source]:

USAMO 2008

</example math problem>

\#\# Summarization

The problem asks for the possible values of $n$ for a regular n-sided polygon that can be completely triangulated into isosceles triangles using non-intersecting diagonals. The solution involves analyzing the properties of the diagonals forming isosceles triangles and deducing that $n$ can be expressed in terms of powers of 2.

\#\# Difficulty

Rating: 7

Reason: The problem involves understanding properties of isosceles triangles in the context of polygon triangulation and requires critical reasoning to establish relationships between the number of sides and powers of 2, making it more complex than typical undergraduate-level problems.

=== report over ===
\\
\\

<math problem>

[QUESTION]:

***Question***

[SOLUTION]:

***Solution***

[SOURCE]:

***SOURCE***

</math problem>
\\
\\

Your answer should be in JSON format for example:

```json

\{

"Rating": YOUR RATING,

"Reason": YOUR JUSTIFICATION,

\}

```

}

\subsection{Topic Domain}
\label{app:prompt_for_topics}
\promptbox{Meta Prompt for Topic Domain Annotation}{
I am a teacher, and I have some high-level Olympiad math problems. \\
I want to categorize the domain of these math problems. \\
\\
\# OBJECTIVE \# \\
1. Summarize the math problem in a brief sentence, describing the concepts involved in the math problem. \\
2. Categorize the math problem into specific mathematical domains. Please provide a classification chain, for example, Applied Mathematics -> Probability -> Combinations. \\
The following is a basic classification framework in the field of mathematics.   \\
<math domains>    \\
... \\
</math domains>

\# STYLE \#

Data report.

\# TONE \#

Professional, scientific.

\# AUDIENCE \#

Students. Enable them to better understand the domain of the problems.

\# RESPONSE: MARKDOWN REPORT \#

\#\# Summarization

[Summarize the math problem in a brief paragraph.]

\#\# Math domains

[Categorize the math problem into specific mathematical domains, including major domains and subdomains.]

\# ATTENTION \#

- The math problem can be categorized into multiple domains, but no more than three. Separate the classification chains with semicolons(;).

- Your classification MUST fall under one of the aforementioned subfields; if it really does not fit, please add "Other" to the corresponding branch. For example: Algebra -> Intermediate Algebra -> Other. Only the LAST NODE is allowed to be "Other"; the preceding nodes must strictly conform to the existing framework.

- The math domain must conform to a format of classification chain, like "Applied Mathematics -> Probability -> Combinations".

- Add "=== report over ===" at the end of the report.
\\
\\
<example math problem>.  

[Question]

Determine the greatest real number $ C $, such that for every positive integer $ n\ge 2 $, there exists $ x_1, x_2,..., x_n \in [-1,1]$, so that $$\prod_{1\le i<j\le n}(x_i-x_j) \ge C^{\frac{n(n-1)}{2}}$$

</example math problem>    

\#\# Summarization

The problem seeks to find the greatest real number $C$ such that, for every integer $n \geq 2$, there exist real numbers $x_1, x_2, \dots, x_n \in [-1,1]$ satisfying the inequality $\prod_{1 \leq i < j \leq n} (x_i - x_j) \geq C^{\frac{n(n-1)}{2}}$. This involves maximizing $C$ to ensure the product of all pairwise differences among $n$ points in the interval $[-1,1]$ is at least $C$ raised to the power of the number of such pairs, $\frac{n(n-1)}{2}$.

\#\# Math domains

Algebra -> Intermediate Algebra -> Inequalities; Discrete Mathematics -> Combinatorics  

=== report over ===
\\
\\
<example math problem>

[Question]

Given integer $n\geq 2$. Find the minimum value of $\lambda {}$, satisfy that for any real numbers $a_1$, $a_2$, $\cdots$, ${a_n}$ and ${b}$,
$$\lambda\sum\limits_{i=1}^n\sqrt{|a_i-b|}+\sqrt{n\left|\sum\limits_{i=1}^na_i\right|}\geqslant\sum\limits_{i=1}^n\sqrt{|a_i|}.$$

</example math problem>   

\#\# Summarization

Let \( n \geq 2 \) be an integer. The problem seeks the minimum value of \( \lambda \) such that for any real numbers \( a_1, a_2, \ldots, a_n \) and \( b \), the inequality \( \lambda \sum_{i=1}^n \sqrt{|a_i - b|} + \sqrt{n \left| \sum_{i=1}^n a_i \right|} \geqslant \sum_{i=1}^n \sqrt{|a_i|} \) holds. The goal is to find the smallest \( \lambda \) that ensures this inequality is satisfied for all possible choices of \( a_i \) and \( b \).  

\#\# Math domains

Algebra -> Intermediate Algebra -> Inequalities; 

Calculus -> Differential Calculus -> Applications of Derivatives.  

=== report over ===
\\
\\

<math problem>

[Question]

***Question***

[Solution]

***Solution***

[Source]

***Source***

</math problem>
\\
\\

Your answer should be in JSON format for example:

"""~json

\{

    "Summary": "YOUR\_SUMMARY",
    
    "Domains": [domain1, ...]

\}

"""~
}

\subsection{Theorem Variant Construction}
\label{sec:appedix-template-theorem-variant}

\promptbox{Meta Prompt for Constructing Theorem Variants}{
I'm going to give you a math proof question and its solution. Your task is to follow the steps below to write a new question based on the given one. Here is the original question and solution:\\

```question

<question>

```,

```solution

<solution>

``` \\

Please follow these steps:

1. The original question asks to "prove or disprove" a statement, where the statement can be "proved". Please write a new question by negating the original statement, so that it can now be "disproved". For example, if the original statement is $x = y$, you may change it to $x < y$ or $x \neq y$; if the statement is "there exists xxx", you may change it to "there does not exist xxx". When negating the original question, you should make minimal changes, i.e. leave as much background information unchanged as possible.\\

2. After changing the question, the solution should be changed accordingly. You do not have to write a new solution, and the original solution can probably be reused. For example, if the original question asks to prove $x = y$ and the new question asks to prove $x < y$, you may simply add a step to the original proof like "since we proved $x = y$, the statement $x < y$ is disproved". However, check the wording of the solution so that it tries to "prove" the statement at first, and then naturally transit to finding that it cannot be proved, but can be disproved instead.\\

3. The original solution ends with "$\backslash$boxed\{proved\}". Your new solution should end with "$\backslash$boxed\{disproved\}".

Output the new question and solution in two blocks:

```question

new question

```

```solution

corresponding solution

```
}

\subsection{System Prompt for RL Training}\label{sec:appedix-system-prompt}

\promptbox{System Prompt for RL Training}{
A conversation between User and Assistant. The User gives a statement, and the Assistant either proves or disproves it. The Assistant first thinks about the reasoning process in the mind and then provides the User with the answer. The reasoning process is enclosed within <think> </think> and the answer is enclosed within <answer> </answer> tags, respectively, i.e., <think> reasoning process here </think> <answer> answer here </answer>. If you prove the statement, answer with "proved". If you disprove the statement, answer with "disproved". You must put your answer inside <answer> </answer> tags, i.e., <answer> $\backslash$boxed\{proved\} or $\backslash$boxed\{disproved\} </answer>. And your final answer will be extracted automatically by the $\backslash$boxed\{\} tag.
}

\subsection{Process Evaluation Framework}\label{sec:appendix-process-evaluation-prompt}
\promptbox{Meta Prompt for Process Evaluation}{
You are an expert in scoring solutions for mathematical proof questions. The following question asks to prove or disprove a statement, where the statement may be either true or false. The test subject is asked to end their proof with $\backslash$boxed\{proved\} if they prove the statement to be true, and $\backslash$boxed\{disproved\} if they prove the statement to be false.\\

The question:

```<question>```\\
The ground truth of the statement:

```<answer>```\\
The test subject's solution:

```~<solution>```\\

Your task is to evaluate the proof's quality and assign a score from 0 to 1 based on four criteria: logical validity (40\%), completeness (30\%), correctness (20\%), and clarity (10\%).

Instructions:

1. Analyze the proof step by step.

2. For each criterion:

   - Logical Validity: Check if each step follows logically from the previous one. Flag any logical errors.

   - Completeness: Verify if all necessary cases and steps are included to prove the theorem.

   - Correctness: Confirm if the final conclusion is correct.

   - Clarity: Assess if the proof is clear, unambiguous, and well-explained.
   
3. Assign a sub-score (0 to 1) for each criterion and compute the total score using the weights: (0.4 × validity) + (0.3 × completeness) + (0.2 × correctness) + (0.1 × clarity).

4. Provide a brief explanation (2-3 sentences) summarizing any errors or issues and justifying the score.

Final output format:

```\\
\{\\
    "score": float,\\
    "validity": float,\\
    "completeness": float,\\
    "correctness": float,\\
    "clarity": float,\\
    "explanation": str\\
\}

```\\
where "score" is the total score, and "validity", "completeness", "correctness", "clarity" are the subscores.
}

\end{document}

%% file: tables/main_results.tex
\begin{table*}
\centering
\resizebox{0.9\linewidth}{!}
{%
\begin{tabular}{llcccccccc} 
\toprule
\multirow{2}{*}{\textbf{Model}} & \multirow{2}{*}{\textbf{Strategy}} & \multicolumn{2}{c}{\textbf{FIMO}} & \multicolumn{2}{c}{\textbf{HMMT}} & \multicolumn{2}{c}{\textbf{Putnam}} & \multicolumn{2}{c}{\textbf{Avg.}} \\ 
\cmidrule(rl){3-4}\cmidrule(rl){5-6}\cmidrule(rl){7-8}\cmidrule(rl){9-10}
 &  & \textit{out.} & \textit{proc.} & \textit{out.} & \textit{proc.} & \textit{out.} & \textit{proc.} & \textit{out.} & \textit{proc.} \\ 
\midrule
\multirow{6}{*}{Qwen2.5-1.5B} & \multicolumn{9}{l}{\textbf{\textit{SFT}}} \\
 & OpenR1-Proof & 20.63 & 8.66 & 11.86 & 4.80 & 35.42 & 18.98 & 22.64 & 10.81 \\
 & DeepTheorem & 31.75 & \textbf{18.86} & 15.25 & 9.41 & 36.46 & 21.43 & 27.82 & \textbf{16.57} \\
 & \multicolumn{9}{l}{\textbf{\textit{RL}}} \\
 & OpenR1-Proof & \textbf{34.92} & 8.54 & 16.95 & 5.10 & \textbf{55.21} & 17.92 & 35.69 & 10.52 \\
 & DeepTheorem & 31.75 & 15.23 & \textbf{23.73} & \textbf{10.15} & 52.08 & \textbf{22.79} & \textbf{35.85} & 16.06 \\ 
\midrule
\multirow{6}{*}{Qwen2.5-3B} & \multicolumn{9}{l}{\textbf{\textit{SFT}}} \\
 & OpenR1-Proof & 23.81 & 12.85 & 15.25 & 6.90 & 43.75 & 27.96 & 27.60 & 15.90 \\
 & DeepTheorem & 33.33 & 20.38 & 20.34 & 12.15 & 36.46 & 25.43 & 30.04 & 19.32 \\
 & \multicolumn{9}{l}{\textbf{\textit{RL}}} \\
 & OpenR1-Proof & 34.92 & 14.33 & 23.73 & 11.72 & \textbf{57.29} & \textbf{35.11} & \textbf{38.65} & 20.39 \\
 & DeepTheorem & \textbf{38.10} & \textbf{23.39} & \textbf{25.42} & \textbf{13.56} & 52.08 & 33.84 & 38.53 & \textbf{23.60} \\ 
\midrule
\multirow{6}{*}{Qwen2.5-7B} & \multicolumn{9}{l}{\textbf{\textit{SFT}}} \\
 & OpenR1-Proof & 30.16 & 18.23 & 15.25 & 8.63 & 48.96 & 32.95 & 31.46 & 19.94 \\
 & DeepTheorem & 34.92 & 26.69 & 22.03 & 15.41 & 41.67 & 33.50 & 32.87 & 25.20 \\
 & \multicolumn{9}{l}{\textbf{\textit{RL}}} \\
 & OpenR1-Proof & 42.86 & 22.79 & 25.42 & 13.15 & \textbf{60.42} & 38.94 & 42.90 & 24.96 \\
 & DeepTheorem & \textbf{55.56} & \textbf{39.07} & \textbf{28.81} & \textbf{20.85} & 57.29 & \textbf{42.20} & \textbf{47.22} & \textbf{34.04} \\
\bottomrule
\end{tabular}
}
\caption{Outcome (out.) and Process (proc.) evaluation of models trained on OpenR1-Math-Proof and \method.}
\label{tab:main-results}
\end{table*}

%% file: tables/other_models.tex
\begin{table*}

\resizebox{0.98\linewidth}{!}{%
\begin{tabular}{lccccccccrr} 
\toprule
\multirow{2}{*}{\textbf{Model }} & \multicolumn{2}{c}{\textbf{FIMO}} & \multicolumn{2}{c}{\textbf{HMMT}} & \multicolumn{2}{c}{\textbf{Putnam}} & \multicolumn{2}{c}{\textbf{Avg.}} & \multicolumn{2}{c}{\textbf{\#Rank}} \\ 
\cmidrule(rl){2-3}\cmidrule(rl){4-5}\cmidrule(rl){6-7}\cmidrule(rl){8-9}\cmidrule(rl){10-11}
 & \textit{out.} & \textit{proc.} & \textit{out.} & \textit{proc.} & \textit{out.} & \textit{proc.} & \textit{out.} & \textit{proc.} & \textit{out.} & \textit{proc.} \\ 
\hline\hline
\multicolumn{11}{c}{\textit{Commercial Models}} \\ 
\midrule
Claude3.7-Sonnet & 34.92 & 26.28 & 13.56 & 8.29 & 45.83 & 38.33 & 31.44 & 24.30 & 9 & 9 \\
Gemini2.5-Pro & 57.14 & 54.06 & 57.63 & 49.82 & 64.58 & 58.75 & 59.78 & 54.21 & 2 & 3 \\
Gemini2.5-Flash & 30.16 & 28.95 & 25.42 & 22.02 & 42.71 & 38.98 & 32.76 & 29.98 & 8 & 6 \\
GPT-4o & 34.92 & 30.70 & 16.95 & 14.59 & 22.92 & 18.88 & 24.93 & 21.39 & 12 & 10 \\
o1-mini & 60.32 & 55.23 & 35.59 & 30.90 & 61.46 & 52.88 & 52.46 & 46.34 & 4 & 4 \\
o1 & 66.67 & 61.00 & 47.46 & 47.30 & 62.50 & 57.55 & 58.88 & 55.28 & 3 & 2 \\
o3-mini & 80.95 & 77.61 & 45.76 & 43.47 & 78.12 & 75.12 & 68.28 & 65.40 & 1 & 1 \\ 
\hline\hline
\multicolumn{11}{c}{\textit{Open Source Models}} \\ 
\midrule
Qwen2.5-Inst-7B & 30.16 & 21.13 & 10.17 & 6.83 & 33.33 & 25.39 & 24.55 & 17.78 & 13 & 12 \\
Qwen2.5-Inst-72B & 49.21 & 37.35 & 13.56 & 9.78 & 48.96 & 41.00 & 37.24 & 29.38 & 6 & 7 \\
Qwen2.5-Math-Inst-7B & 28.57 & 18.86 & 3.39 & 1.61 & 25.00 & 18.79 & 18.99 & 13.09 & 16 & 17 \\
Qwen2.5-Math-Inst-72B & 47.62 & 36.02 & 11.86 & 8.61 & 47.92 & 38.04 & 35.80 & 27.56 & 7 & 8 \\
DS-Prover-v1.5-RL-7B & 25.40 & 13.81 & 11.86 & 6.32 & 34.38 & 22.42 & 23.88 & 14.18 & 14 & 16 \\
DS-Prover-v2-7B & 30.16 & 21.86 & 5.08 & 1.71 & 40.62 & 28.54 & 25.29 & 17.37 & 11 & 13 \\
R1-Distill-7B & 6.35 & 4.27 & 0 & 0 & 4.17 & 2.58 & 3.51 & 2.28 & 18 & 18 \\
R1-Distill-70B & 17.46 & 14.05 & 16.95 & 13.52 & 30.21 & 23.10 & 21.54 & 16.89 & 15 & 14 \\
QwQ-32B & 17.46 & 15.41 & 11.86 & 10.10 & 25.00 & 18.19 & 18.11 & 14.57 & 17 & 15 \\
Llama3.3-Inst-70B & 41.27 & 27.33 & 10.17 & 4.12 & 36.46 & 25.30 & 29.30 & 18.92 & 10 & 11 \\
*\method-RL-7B & 55.56 & 39.07 & 28.81 & 20.85 & 57.29 & 42.20 & 47.22 & 34.04 & 5 & 5 \\
\bottomrule
\end{tabular}%
}
\centering
\caption{Evaluation comparison of our model trained with RL on \method~with SOTA LLMs for both commercial models and open source models. *: denotes our method. Inst: Instruct; DS: Deepseek.}
\label{tab:sota-llm-results}
\end{table*}

%% file: sections/5-related-work.tex
\section{Related Work}


\paragraph{Theorem proving}
Theorem proving is a non-trivial task for natural language processing. In the context of LLMs, theorem proving is typically achieved via formal languages such as Lean, Coq, and Isabelle~\citep{zheng2021minif2f,liu2023fimo,tsoukalas2024putnambenchevaluatingneuraltheoremprovers}. LLMs specialized at theorem proving have been proposed over the years, represented by GPT-f~\citep{2020GPT-f}, DeepSeek-Prover~\citep{xin2024deepseek,2024DeepSeek-Prover-V1.5,2025DeepSeek-Prover-V2}, TheoremLlama~\citep{2024TheoremLlama}, InterLM-StepProver~\citep{2024-interlm2.5-stepprover}, and MPS-Prover~\citep{liang2025mpsproveradvancingstepwisetheorem}. Although most of these works focus on formal theorem proving, NaturalProofs~\citep{2021NaturalProofs} and NaturalProver~\citep{2022NaturalProver} have emerged as some of the few works that attend to informal theorem proving. However, NaturalProofs use a domain-specific language from the ProofWiki website. 
In comparison, \methodit~represents the first attempt at exploring the more commonly used LaTeX-based natural language theorem proving at scale.



\paragraph{Learning to reason with RL-zero}  

RL-Zero~\citep{guo2025deepseek} is a streamlined framework designed to develop reinforcement learning (RL) capabilities in large language models (LLMs) without SFT. 
While recent advances in LLM reasoning have been significantly influenced by RL techniques~\citep{jaech2024openai, guo2025deepseek, Team_2024a, grok2025, googlecloud2025}, existing approaches predominantly focus on closed-form questions, addressing only a subset of reasoning problems.

In this work, we investigate the application of RL-Zero in process-oriented reasoning, specifically in informal mathematical theorem-proving tasks. 
To the best of our knowledge, this is the first study to apply the RL-Zero framework to informal theorem proving, marking a significant advance in enabling LLMs to address more diversified reasoning tasks in mathematical and logical domains.

%% file: sections/6-conclusion.tex
\section{Conclusion}



In this paper, we introduced \method, a novel informal theorem-proving paradigm involving a large-scale annotated dataset of 121K IMO-level informal mathematical theorems and corresponding high-quality natural-language proofs, alongside systematically constructed verifiable theorem variants. We further proposed a novel RL-Zero method specifically tailored for informal mathematical reasoning, significantly surpassing supervised fine-tuning frameworks in theorem proving. Comprehensive evaluations involving outcome accuracy and detailed reasoning process assessment demonstrate the effectiveness of our approach, achieving state-of-the-art theorem-proving performance and significantly pushing LLM reasoning boundaries. Through these contributions, \method{} provides a robust foundation for future advancements in automated mathematical theorem proving, leveraging natural language flexibility to empower scalable, human-like reasoning abilities in large language models.